\documentclass[journal]{IEEEtran}
\usepackage[pdftex]{graphicx}
\usepackage[colorlinks,linkcolor=blue]{hyperref}
\usepackage{amsmath}
\usepackage{amsfonts,amssymb}
\usepackage{algorithmic}
\usepackage{multirow}
\usepackage{graphicx}
\usepackage{cite}
\usepackage{color}
\usepackage{flushend}
\ifCLASSOPTIONcompsoc
\else
\fi
\begin{document}
%
% paper title
% Titles are generally capitalized except for words such as a, an, and, as,
% at, but, by, for, in, nor, of, on, or, the, to and up, which are usually
% not capitalized unless they are the first or last word of the title.
% Linebreaks \\ can be used within to get better formatting as desired.
% Do not put math or special symbols in the title.
\title{Efficient and Accurate Multi-scale Topological Network for Single Image Dehazing}
%
%
% author names and IEEE memberships
% note positions of commas and nonbreaking spaces ( ~ ) LaTeX will not break
% a structure at a ~ sfo this keeps an author's name from being broken across
% two lines.
% use \thanks{} to gain access to the first footnote area
% a separate \thanks must be used for each paragraph as LaTeX2e's \thanks
% was not built to handle multiple paragraphs
%

\author{Qiaosi Yi\textsuperscript{$\dagger$}, Juncheng Li\textsuperscript{$\dagger$}, Faming Fang\textsuperscript{*}, Aiwen Jiang, Guixu Zhang% <-this % stops a space
\thanks{$\dagger$: Equal contribution. *: The corresponding author.}% <-this % stops a space
\thanks{Q. Yi, J. Li, F. Fang, and G. Zhang are with the Shanghai Key Laboratory of Multidimensional Information Processing, East China Normal University, Shanghai, China, and also with the school of Computer Science and Technology, East China Normal University, Shanghai, China. (E-mail: qiaosiyijoyies@gmail.com, cvjunchengli@gmail.com, fmfang@cs.ecnu.edu.cn, gxzhang@cs.ecnu.edu.cn)}
\thanks{A. Jiang is with the School of Computer and Information Engineering, Jiangxi Normal University, Nanchang, China. (E-mail: jiangaiwen@jxnu.edu.cn)}
}

% note the % following the last \IEEEmembership and also \thanks - 
% these prevent an unwanted space from occurring between the last author name
% and the end of the author line. i.e., if you had this:
% 
% \author{....lastname \thanks{...} \thanks{...} }
%                     ^------------^------------^----Do not want these spaces!
%
% a space would be appended to the last name and could cause every name on that
% line to be shifted left slightly. This is one of those "LaTeX things". For
% instance, "\textbf{A} \textbf{B}" will typeset as "A B" not "AB". To get
% "AB" then you have to do: "\textbf{A}\textbf{B}"
% \thanks is no different in this regard, so shield the last } of each \thanks
% that ends a line with a % and do not let a space in before the next \thanks.
% Spaces after \IEEEmembership other than the last one are OK (and needed) as
% you are supposed to have spaces between the names. For what it is worth,
% this is a minor point as most people would not even notice if the said evil
% space somehow managed to creep in.

% The paper headers
\markboth{JOURNAL OF LATEX CLASS FILES, VOL. 14, NO. 8, AUGUST 2015 1
}%
{Shell \MakeLowercase{\textit{et al.}}: Bare Demo of IEEEtran.cls for IEEE Journals}
% The only time the second header will appear is for the odd numbered pages
% after the title page when using the twoside option.
% 
% *** Note that you probably will NOT want to include the author's ***
% *** name in the headers of peer review papers.                   ***
% You can use \ifCLASSOPTIONpeerreview for conditional compilation here if
% you desire.

% If you want to put a publisher's ID mark on the page you can do it like
% this:
%\IEEEpubid{0000--0000/00\$00.00~\copyright~2015 IEEE}
% Remember, if you use this you must call \IEEEpubidadjcol in the second
% column for its text to clear the IEEEpubid mark.

% use for special paper notices
%\IEEEspecialpapernotice{(Invited Paper)}

% make the title area
\maketitle

% As a general rule, do not put math, special symbols or citations
% in the abstract or keywords.
\begin{abstract}
Single image dehazing is a challenging ill-posed problem that has drawn significant attention in the last few years.
Recently, convolutional neural networks have achieved great success in image dehazing. 
However, it is still difficult for these increasingly complex models to recover accurate details from the hazy image.
In this paper, we pay attention to the feature extraction and utilization of the input image itself.
To achieve this, we propose a Multi-scale Topological Network (MSTN) to fully explore the features at different scales.
Meanwhile, we design a Multi-scale Feature Fusion Module (MFFM) and an Adaptive Feature Selection Module (AFSM) to achieve the selection and fusion of features at different scales, so as to achieve progressive image dehazing.
This topological network provides a large number of search paths that enable the network to extract abundant image features as well as strong fault tolerance and robustness.
In addition, ASFM and MFFM can adaptively select important features and ignore interference information when fusing different scale representations.
Extensive experiments are conducted to demonstrate the superiority of our method compared with state-of-the-art methods.
\end{abstract}

% Note that keywords are not normally used for peerreview papers.
\begin{IEEEkeywords}
Image dehazing, multi-scale topological network, feature fusion, adaptive feature selection.
\end{IEEEkeywords}

% For peer review papers, you can put extra information on the cover
% page as needed:
% \ifCLASSOPTIONpeerreview
% \begin{center} \bfseries EDICS Category: 3-BBND \end{center}
% \fi
%
% For peerreview papers, this IEEEtran command inserts a page break and
% creates the second title. It will be ignored for other modes.
\IEEEpeerreviewmaketitle

\section{Introduction}
\IEEEPARstart{H}{aze} is a common atmospheric phenomenon produced by small floating particles.
Particulate matter floating in the air causes light scattering and attenuation, thereby reducing the visibility of distant objects.
However, hazy images will cause difficulties with their processing and analysis, which will seriously affect the performance of downstream tasks such as image classification, image segmentation, object detection, crowd counting, and other high-level computer vision tasks.
This is not conducive to the construction of safe and stable artificial intelligence systems, such as video surveillance systems and unmanned driving systems.
In order to solve this problem, the task of image dehazing, especially single image dehazing came into being and has drawn significant attention in the last few years.

\begin{figure}
	\centering
	\includegraphics[scale=0.78]{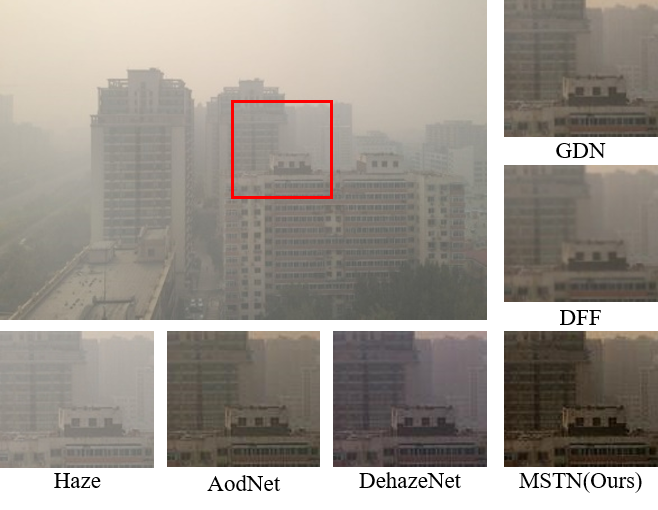}
	\caption{An example of single image dehazing. Obviously, the haze-free image reconstructed by our MSTN shows better visual effect.}
	\label{head}
\end{figure}

Single image dehazing is an extremely hot topic in computer vision, which aims to reconstruct a haze-free image from the hazy one (Fig.~\ref{head}).
However, due to the absorption and reflection of the haze, the captured scene image will suffer from color distortion, blur, and low contrast, which causes the quality of the image to deteriorate.
Therefore, single image dehazing still is a challenging task and many methods have been proposed to try to solve this task.

The atmosphere scattering model provides a theoretical basis for hazy imaging and is also a basic method for image dehazing. 
As shown in Eq.~(\ref{Eq.1}), the atmosphere scattering model can be defined as:
\begin{equation} \label{Eq.1}
	\small
	I_{i}(x)=J_{i}(x)t(x)+A(1-t(x)), \quad i=1,2,3
\end{equation}
where $I(x)$ is the observed hazy image, $J(x)$ is the clear image, and $A$ represents the global atmospheric light intensity.
Meanwhile, $i$ denotes ${R,G,B}$ channels in a RGB image, $x$ represents the pixel locations,
$I_{i}(x)$ and $J_{i}(x)$ represents the value of the $i$-th channel of the hazy or clear image in location $x$.
$t(\cdot)$ is the medium transmission function, $t(x)=e^{-\beta d(x)}$, $\beta$ and $d(x)$ represent the atmosphere scattering parameter and the scene depth, respectively.
The atmosphere scattering model shows that single image dehazing is an ill-posed problem, which is a challenging task without the priors of $A$ and $t(x)$
\begin{equation} \label{Eq.2}
	\small
	J(x)=\frac{I(x)-A}{t(x)}+A.
\end{equation}

In the past, in order to better deal with the problem of single dehazing, many methods have been proposed to learned different prior knowledge to estimate $A$ and $t(x)$, then obtain the haze-free image based on the atmosphere scattering model. 
For example, dark channel prior~\cite{5206515}, color attenuation prior\cite{zhu2015fast}, and non-local prior~\cite{berman2016non} are proposed for transmission function $t(\cdot)$ estimation.
Meanwhile, some works focus on estimating the atmospheric light $A$, such as~\cite{sulami2014automatic, wang2017fast}.
Based on the estimated $\widehat{t}(x)$ and $\widehat{A}$, the clear image $\widehat{J}$ can be recovered by the following formulation
\begin{equation} \label{Eq.3}
	\small
	J(x)=\frac{I(x)-\widehat{A} \cdot (1 - \widehat{t}(x))}{\widehat{t}(x)} = \frac{1}{\widehat{t}(x)} I(x) - \frac{\widehat{A}}{\widehat{t}(x)} + \widehat{A}.
\end{equation}
However, due to the complexity of the real environment, the prior may be easily violated in practice.
Therefore, the methods based on the atmosphere scattering model may not be able to accurately estimate the transmission map and the global atmospheric light intensity, resulting in the inability to obtain clear haze-free images.
This will greatly limit the model speed, versatility, and performance.

Recently, convolutional neural networks (CNNs) have achieved remarkable success in many computer vision tasks and also greatly promoted the development of image dehazing.
With the powerful feature extraction capabilities of CNN, more and more CNN-based image dehazing methods have been proposed for $A$ and $t(x)$ estimation or directly learn the mapping between hazy and clear images.
For example, Cai et al. proposed the first CNN model (Dehazenet~\cite{cai2016dehazenet}) to directly remove haze from the hazy image.
Li et al. proposed a all-in-one dehazing Network (AODNet~\cite{li2017aod}), which based on a re-formulated atmospheric scattering model and directly generates the clean image through a light-weight CNN.
After that, CNN-based image dehazing models have been blooming and refreshing the best results, including PFF-Net~\cite{mei2018progressive}, DCPDN~\cite{zhang2018densely}, EPDN~\cite{qu2019enhanced}, PDR-Net~\cite{8792133}, GDN~\cite{liu2019griddehazenet}, and DFF~\cite{dong2020multi}.
Although the aforementioned methods have made a big breakthrough in image dehazing.
However, most existing image dehazing models have the following shortcomeing:

\begin{enumerate}
	\item Most existing methods focus on microstructure design, that is, build the network and achieve image dehazing by stacking the carefully designed feature extraction modules.
	This modular design strategy ignores the connectivity of the model.
	In addition, this cascaded structural design greatly reduces the possible topological paths, which is not conducive to building an effective model.
	\item Most existing methods ignore the morphological difference of hazy images at different scales. 
	Therefore, these models do not pay attention to the extraction, propagation, fusion, and utilization of the multi-scale image features.
	\item The structure of these models is getting bigger, deeper, and more complex, which is not conducive to building an efficient and real-time dehazing model.
\end{enumerate}

According to~\cite{lin2017feature}, the lower-level features have higher resolution and more texture details but lower semantic information, and the higher-level features have more semantic information but fewer texture details. 
Therefore, the core of this work is to build an effective network that can fully extract and utilize image features at different stages.
Specifically, we propose a Multi-scale Topological Network (MSTN) to progressive remove the haze in the hazy image.
MSTN is a topological network that can promote the transmission and utilization of image feature flows.
Meanwhile, these topological sub-nets enable the network to detect rich image features while increasing the fault tolerance of the model.
Considering the effectiveness of multi-scale image features, we also introduce the multi-scale strategy into the model.
Therefore, MSTN can be considered as a multi-branch network and each branch is used to extract images features at different scales.
However, if these branches are independent of each other, they cannot form a topological network, which will greatly reduce the model performance. 
In order to solve this problem, we take the output of the lower-resolution branch as the input of the previous branch.
In addition, we design an Adaptive Feature Selection Module (AFSM) and a Multi-scale Feature Fusion Module (MFFM) to realize automatic selection and fusion of multi-scale image features, which helps to make full use of the features of the image itself.

In summary, the main contributions of this work include:
\begin{itemize}
	\item We reveal the importance of topology for deep network design and proposed a Multi-scale Topological Network (MSTN) for image dehazing, which shows stronger robustness and versatility.
	Compared with existing models, MSTN achieves better results with less execution time.
	\item We design a Multi-scale Feature Fusion Module (MFFM), which can promote the interaction and fusion between different scale features, thereby improving the utilization of multi-scale features.
	\item We propose an Adaptive Feature Selection Module (AFSM) to automatically select image features at different scales.
	Compared with directly adding all different scales features together, this module can effectively remove redundant features to achieve better feature extraction and utilization.
\end{itemize}
The remaining parts of this paper are organized as follows. Section II reviews related works including prior-based and learning-based image dehazing methods, topological network, multi-scale feature extraction, and Attention Mechanism. A detailed explanation of
the proposed MSTN is given in Section III. The experimental
results and ablation analysis are presented in Section IV and V,
respectively. Finally, we draw a conclusion in Section VI.
\section{Related Work}
\subsection{Single Image Dehazing}
\subsubsection{Prior-based Methods}
The prior-based methods use the characteristics of the image to estimate $A$ or $t(x)$ and recover the clear haze-free image according to the atmospheric scattering model.
For example, Fattle~\cite{fattal2008single} added a surface shadow factor to the atmospheric scattering model to estimate the transmission map; 
He et al.~\cite{5206515} proposed a dehazing algorithm based on dark channels prior (DCP), which estimates the transmission map by the DCP;
Fattle~\cite{fattal2014dehazing} proposed a color-line prior dehazing method based on the observation that the color of a small image patch exhibits a one-dimensional distribution in the RGB color space;
Although these prior-based methods have achieved varying degrees of success, their performance depends on the accuracy and validity of the proposed priors.

\begin{figure*}
	\begin{center}
		\includegraphics[scale=0.45]{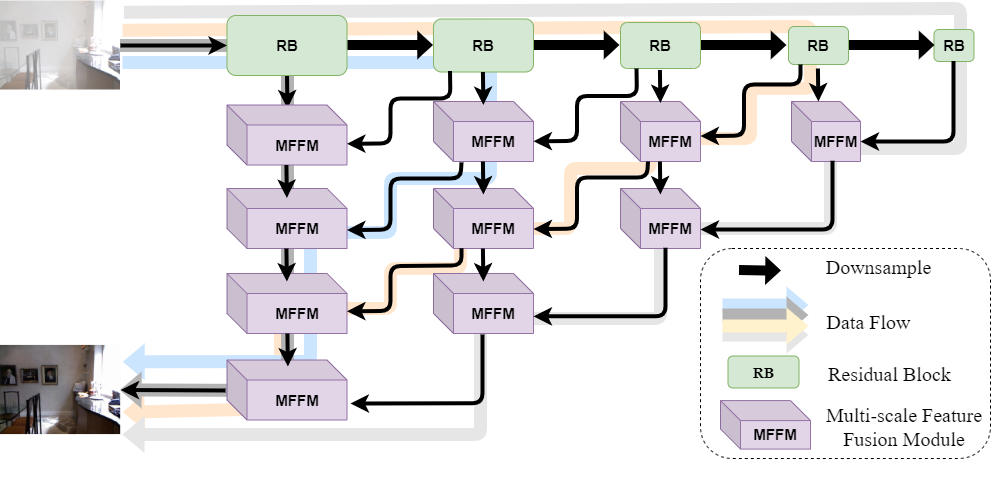}
	\end{center}
	\caption{The architecture of our proposed Multi-scale Topological Network (MSTN), which is a multi-branch network that contains five rows and five columns.}
	\label{MSTN}
\end{figure*}

\subsubsection{Learning-based Methods}
With the rise of CNN, learning-based methods have become the mainstream method and have achieved tremendous development.
These methods usually use CNN to estimate $A$ and $t(x)$ or directly recover the clear haze-free image by an efficient CNN.
For example, Cai et al.~\cite{cai2016dehazenet} adopt a three-layer convolutional neural network (Dehazenet) to estimate the transmission map; 
Zhang et al.~\cite{zhang2018densely} proposed a density dehazing network (DCPDN) that can simultaneously estimate the transmission map and atmospheric light intensity;
Liu et al.~\cite{liu2019griddehazenet} proposed a grid network (GDN) to directly reconstruct clear images;
Dong et al.~\cite{dong2020multi} proposed multi-scale based deep network which works on strengthen-operate-subtract-boosting strategy for image dehazing (DFF).
Although these learning-based methods have made great progress, they did not fully extract the features of the hazy image itself, resulting in sub-optimal reconstruction results.

\subsubsection{Topological Network}
Topology is a discipline that studies the properties of geometric figures or spaces that can remain unchanged after continuously changing shapes. 
It only considers the positional relationship between objects without considering their shape and size. 
The biggest characteristic of the topological architecture is its invariance under local deformation, which can simplify the network design.
For example, Attara et al.~\cite{attar2017classification} proposed a method that based on supervised machine learning algorithms and utilizes the topological similarities of networks for the classification task. 
Li et al.~\cite{li2019lightweight} proposed a recursive fractal network that can construct an infinite variety of topological structures through a simple basic component.
At the same time, this structure has been proven to be more fault-tolerant, stable, and robust. 
Thus, the topological structure will be the focus of our research.

\subsubsection{Multi-scale Feature Extraction and Utilization}
Plenty of researches have pointed out that the image will exhibit different characteristics at different scales.
Therefore, making full use of the features of the input image itself can further improve model performance.
In recent years, many works have been proposed for multi-scale features extraction and utilization, which can be roughly divided into two categories:
(i) The most widely used method is to obtain images with different resolutions after multiple downsampling operations, and then extract the features separately. 
This type of method is commonly used in image segmentation and object detection tasks, such as FPN~\cite{lin2017feature}, FPT~\cite{zhang2020feature}, and PyConvResNet~\cite{duta2020pyramidal}.
(ii) Another method is to extract image features by different convolutional kernels.
This type of method adjusts the size of the receptive field through different convolutional kernels, so as to achieve multi-scale feature extraction.
The most famous methods includs VGG~\cite{szegedy2015going}, MSRN~\cite{li2018multi}, MSIN~\cite{li2019multi}, MSFFRN~\cite{qin2020multi}, and MDCN~\cite{li2020mdcn}.
In this work, we aim to introduce the multi-scale strategy into the topological network to better mine and utilize multi-scale image features.

\subsubsection{Attention Mechanism}
Recent years, the attention mechanism~\cite{hu2018squeeze, woo2018cbam, li2019selective, 8584101, 8424031, 9186194} has been shown significant advantages in a range of tasks, from neural machine translation in natural language processing to image captioning in image understanding. The important information is highlighted by the attention mechiam and the less useful information is suppressed. Attention has been widely used in recent applications such as person Re-ID, image recovery, and segmentation. To boost the performance of image classification, SENet~\cite{hu2018squeeze} brings an effective, lightweight gating mechanism to self-recalibrate the feature map via channel-wise importances. Beyond channel, CBAM~\cite{woo2018cbam} introduce spatial attention in a similar way. Furthermore, SKNet~\cite{li2019selective} focus on the adaptive receptive field size of neurons by introducing the attention mechanisms. Different them, the core of our method, which takes different scales features as input for learning and output the selected and fused features, is to automatically respond and select features from different scale inputs. In the image dehazing task, the GDN~\cite{liu2019griddehazenet} learn a coefficient for adding different scale feature as the attention mechanism. Moreover, the coefficient is learned by global learning and do not depend on the different scale feature. But, our method utilize the attention mechanism to learn the relationship between the different scale feature and highlight the most informative feature expressions in the different scale feature.

\section{Multi-scale Topological Network (MSTN)}
In this paper, we propose a Multi-scale Topological Network (MSTN) for single image dehazing.
As shown in Fig.~\ref{MSTN}, MSTN is essentially a multi-branch network, which contains $i$ rows and $j$ columns.
Among them, $i$ denotes the depth of the network and $j$ represents the scale of the model, respectively.  
Meanwhile, we can clearly observe that at each branch the model contains one Residual Block (RB~\cite{he2016deep}) and several Multi-scale Feature Fusion Module (MFFM).
It is worth noting that all these branches are used to extract image features at different scales and the input of each branch is obtained from the output of the previous branch through downsample operation. 
In addition, RBs are used for feature extraction and MFFMs are the core module of MSTN, which are used for multi-scale feature selection and fusion.
However, if these branches are independent of each other, multi-scale features cannot interact together, which will greatly reduce the model performance. 
In order to solve this problem, we introduce skip connection between the adjacent branches.
In other words, the outputs of the current branch are sent to the previous branch.
Therefore, image features at different scales can be transferred, interacted, and merged via the MFFM.

Define $I_{hazy}$ and $I_{clear}$ as the input hazy image and the reconstructed haze-free image, the model can be defined as 
\begin{equation} \label{Eq.4}
	\small
	I_{clear} = F_{MSTN}(I_{hazy}),
\end{equation}
where $F_{MSTN}(\cdot)$ represents the proposed MSTN.
As mentioned above, MSTN is a multi-branch network that contains $i$ rows and $j$ columns, each row denotes the depth of the network and each column denotes the different scales of the model.
We define the first row and first column as $i=0$ and $j=0$, respectively.
Therefore, the outputs ($R_{i,j}$) of each RB or MFFM can be defined as

\begin{equation}\label{Eq.5}
	\small
	R_{i,j} = \left\{
	\begin{aligned}
		& F_{i,j}(I_{hazy}) &when\; i=0, j=0 \\
		& F_{i,j}(R^{'}_{i,j-1}) &when\; i=0, j>0 \\
		& M_{i,j}(R_{i-1,j},R_{i-1,j+1}) &when\; i>0
	\end{aligned}
	\right.
\end{equation}
where $F_{i,j}(\cdot)$ and $M_{i,j}(\cdot)$ denote the operation of RB and MFFM in the $i$-th row and $j$-th column, respectively. 
Meanwhile, $R^{'}$ is the result obtained by the dowsample operation
\begin{equation} \label{Eq.6}
	\small
	R^{'} = R \downarrow_{2}.
\end{equation}
It is worth noting that the downsampling operation is realized using a convolutional layer instead of traditional methods such as bilinear or bicubic interpolation.
This is because bilinear and bicubic will cause a lot of information to be lost, which is not conducive to image reconstruction, so we use convolutional layer to let the model automatically learn the redundant features that need to be removed.

During training, MSTN is optimized with $L_{1}$ loss function.
Therefore, given a training dataset $\left \{ I_{hazy}^{n}, I_{clear}^{n}  \right \}_{n}^{N}$, we solve
\begin{equation} \label{Eq.7}
	\small
	\hat{\theta} = arg\,\min\limits_{\theta}\, \frac{1}{N}\sum_{n=1}^{N}  \left \| F_{\theta}(I_{hazy}^{n}) -  I_{clear}^{n} \right \|_{1},
\end{equation}
where $\theta$ denotes the parameter set of our model and $F(\cdot)$ represents the proposed MSTN.
Each module of the network will be described in the following sections.

\subsection{Multi-scale Topological Architecture}
In this paper, we propose a multi-scale topological architecture as the backbone of MSTN.
Similar to the Feature Pyramid Network~\cite{lin2017feature}, MSTN also adopts the pyramid-like structure to obtained multi-scale image features.
In other words, we gradually downsample the resolution of the image and extract image features at different scales.
After that, we progressive restore the resolution of the image and use the extracted multi-scale features to reconstruct the final haze-free image.
This strategy can fully mine the potential features of the input image itself, improve the model performance, and reduce the memory consumed during runtime.
However, most of the previous works are simply add all the features extracted from each scale branch, which is not conducive to the interaction between different scales features.
In order to solve this problem, we introduce skip connection between the adjacent branches.
Therefore, image features with different scales can be interacted and merged through MFFM.
It is worth noting that the intermediate results of each branch are sent to the corresponding position of the previous branch for feature selection and fusion rather than just the final output.
This allows the hierarchical features to be fully utilized, which can further improve the quality of reconstructed images.
Meanwhile, these skip connections make the model constitute a topological network, which  provide a large number of search paths that enable the network to extract abundant image features to reconstruct high-quality haze-free images.
In Fig.~\ref{MSTN}, the color data flows represent some examples of the path of the model.
Among them, the gray one represents the complete pathway, the dark gray represents the path of the model without multi-scale strategy.
In addition, the blue and orange lines represent two intermediate paths.
This topological architecture makes the network contains multiple sub-networks and all these sub-networks complement each other, greatly improving the stability and fault tolerance of the model.

\begin{figure}
	\begin{center}
		\includegraphics[width=0.95\linewidth]{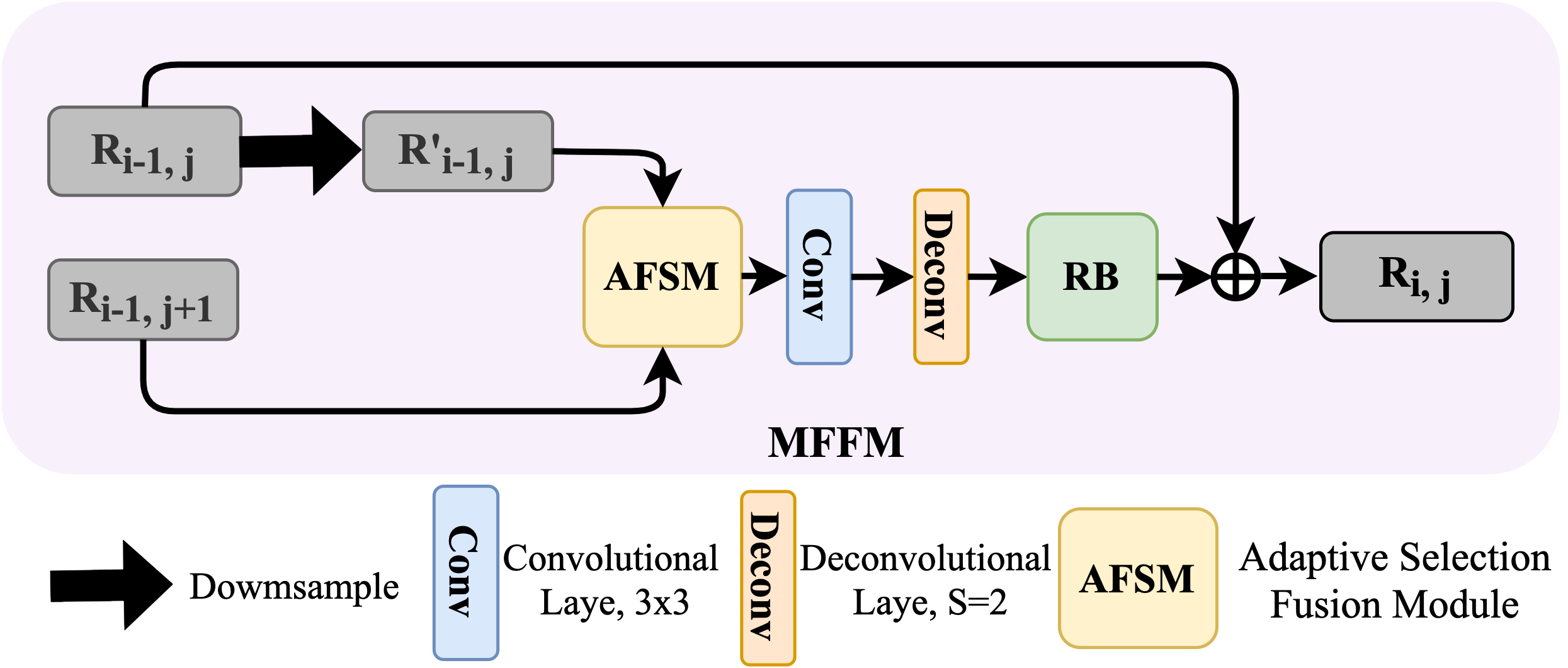}
	\end{center}
	\caption{The complete architecture of MFFM.}
	\label{MFFM}
\end{figure}

\subsection{Multi-scale Feature Fusion Module (MFFM)}
MFFM is the core module of MSTN, which is designed for multi-scale image feature selection, interaction, and fusion.
As shown in Fig.~\ref{MFFM}, MFFM takes $R_{i-1, j}$ and $R_{i-1, j+1}$ as inputs and output the fused image features $R_{i, j}$.
According to the figure, we can clearly observe that the module contains a adaptive feature selection module (AFSM), a convolutional layer, a deconvolutional layer, a residual block (RB), and a residual skip connection.
Firstly, we apply downsampling operation on $R_{i-1, j}$ to obtain $R'_{i-1, j}$.
Then, $R'_{i-1, j}$ and the extracted features from the next branch $R_{i-1, j+1}$ are send to the AFSM for feature selection.
This is a crucially important step that used to automatically select useful features and remove redundant features.
After that, a convolutional layer, a deconvolutional layer, and a residual block are applied to the selected multi-scale image features to obtain new representations. 
Finally, we introduce local residual learning strategy to further improve the information flow.
The introduced residual learning strategy make the module only needs to learn the different areas between the input and output features, which can greatly accelerate the learning process.
Meanwhile, this allows the ASFM to selectively select the missing features from different scale branches.
With the help of MFFM, the model has enough flexibility for selecting important features from different scale representations and can expand the representation capabilities of CNN.

\subsection{Adaptive Feature Selection Module (AFSM)}
\begin{figure}
	\begin{center}
		\includegraphics[width=0.95\linewidth]{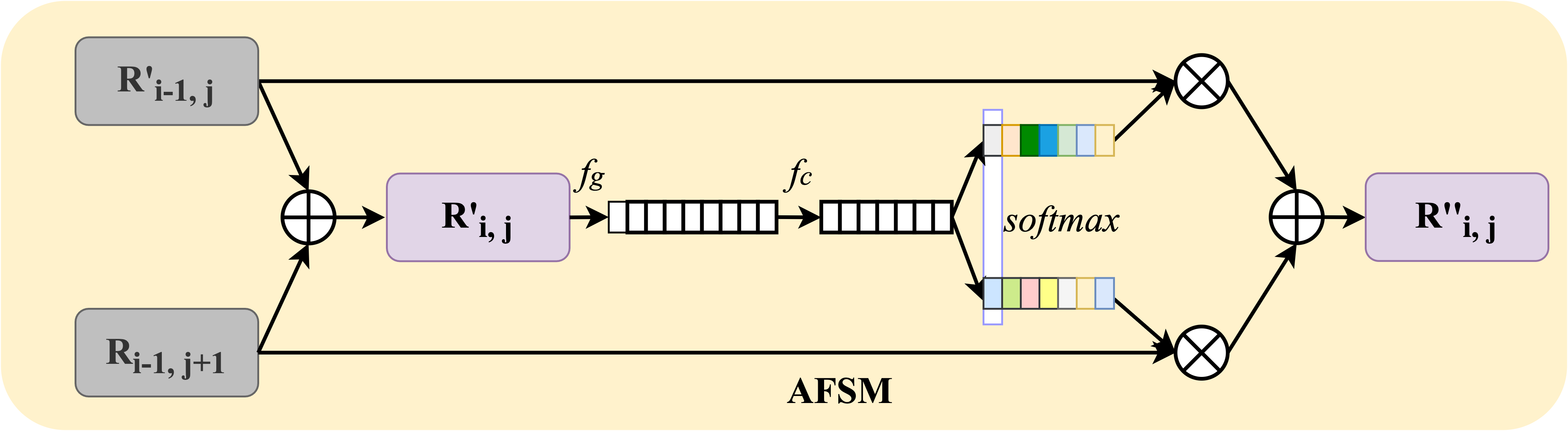}
	\end{center}
	\caption{The complete architecture of AFSM.}
	\label{AFSM}
\end{figure}

According to Lin et al.~\cite{lin2017feature}, we know that image features with different scales have different semantic information. 
Making full use of multi-scale image features can effectively improve the quality of reconstructed images.
However, most existing methods directly cascade or add all multi-scale image features for image reconstruction, it will bring a lot of redundant features that not conducive to building a efficient and accurate model.
In 2019, Li et al.~\cite{li2019selective} proposed a Selective Kernel Networks (SKNet), which can adaptively adjust its receptive field size based on multiple scales of input information.
Inspired by this, we introduce the attention mechanism to the model and propose an Adaptive Feature Selection Module (AFSM) for different image feature selection and fusion.
As shown in Fig.~\ref{AFSM}, AFSM takes different scales features $R^{'}_{i-1, j}$ and $R_{i-1, j+1}$ as inputs for learning, and output the selected and fused $R^{''}_{i, j}$.
Specifically, we first fuse the results from different branches via an element-wise summation:
\begin{equation} \label{Eq.8}
	\small
	R^{'}_{i, j} = R^{'}_{i-1, j} + R_{i-1, j+1}.
\end{equation}

Then we generate channel-wise statistics $\mathbf{s} \in \mathbb{R}^{C}$ by using global average pooling. 
The $c$-th element of $\mathbf{s}$ is calculated by shrinking $R^{'}_{i, j}$ through spatial dimensions $H \times W$:
\begin{equation} \label{Eq.9}
	\small
	\mathbf{s}^{c} = f_{g}(R^{'c}_{i, j})=\frac{1}{H \times W}\sum_{x=1}^{H} \sum_{y=1}^{W} R^{'c}_{i, j}(x, y)
\end{equation}

After that, we applied a fully connected layer to generate compressed features $\mathbf{z} \in \mathbb{R}^{d \times 1}$ for precise and adaptive selection
\begin{equation} \label{Eq.10}
	\small
	\mathbf{z} = f_{c}(\mathbf{s})
\end{equation}

Finally, a soft attention across channels is used to adaptively select important information from different branches.
\begin{equation} \label{Eq.11}
	\small
	a=\frac{e^{\mathbf{A} \mathbf{z}}}{e^{\mathbf{A} \mathbf{z}}+e^{\mathbf{B} \mathbf{z}}}, b=\frac{e^{\mathbf{B} \mathbf{z}}}{e^{\mathbf{A} \mathbf{z}}+e^{\mathbf{B} \mathbf{z}}},
\end{equation}
where $\mathbf{A},\mathbf{B} \in \mathbb{R}^{C\times d}$, and $a$, $b$ denote the attention vector of $R^{'}_{i-1, j}$ and $R_{i-1, j+1}$, respectively. 
Specifically, the softmax function is used on $a$ and $b$, so $a+b=1$. 
After getting $a$ and $b$, the output $R^{''}_{i,j}$ can be calculated as follow:
\begin{equation} \label{Eq.12}
	\small
	{R^{''}_{i,j}} = a \cdot R^{'}_{i-1, j} + b \cdot R_{i-1, j+1}.
\end{equation}

With the help of this module, our MSTN can efficiently and automatically select and fuse multi-scale image features.
This provides a new solution for image restoration task which based on multi-scale architecture.

\begin{table*}[t]
 \caption{Quantitative (PSNR/SSIM) comparisons with SOTA image dehazing methods on RESIDE-SOTS~\cite{li2018benchmarking} (indoor and outdoor) and RESIDE-HSTS~\cite{li2018benchmarking} (Synthetic). The best and second best results are highlighted with \textcolor{red}{red} and \textcolor{blue}{blue} fonts, respectively.}
 \centering
 \renewcommand{\arraystretch}{1.3}
 \setlength{\tabcolsep}{0.4mm}
 \begin{tabular}{|c|c|c|c|c|c|c|c|c|c|c|c|c|}
  \hline
  \multicolumn{2}{|c|}{Method}  & DCP~\cite{5206515} & CAP~\cite{zhu2015fast}   & DehazeNet~\cite{cai2016dehazenet} & MSCNN~\cite{ren2016single} & NLD~\cite{berman2016non} & AODNet~\cite{li2017aod} & DCPDN~\cite{zhang2018densely} & GFN~\cite{ren2018gated}    & GDN~\cite{liu2019griddehazenet}    & DFF~\cite{dong2020multi}   & MSTN (Ours)   \\
  \hline
  \hline
  \multirow{3}{*}{SOTS (Indoor)} & PSNR$\uparrow$  & 16.61 & 19.05 & 21.14 & 17.12 & 17.29 & 19.06   & 15.85 & 22.30  & 32.16 & \textcolor{blue}{33.75} &  \textcolor{red}{35.37} \\
  & SSIM$\uparrow$  & 0.855 & 0.836 & 0.847 & 0.796 & 0.778 & 0.850   & 0.818& 0.880& 0.984  & \textcolor{blue}{0.985} &  \textcolor{red}{0.987}  \\ 
  \hline
  \hline
  \multirow{3}{*}{SOTS (Outdoor)}& PSNR$\uparrow$  & 19.13 & 18.12 & 22.46 & 19.48 & 17.97 & 20.29   & 19.93 & 21.55 & 30.86 & \textcolor{blue}{32.21} &  \textcolor{red}{32.61}   \\
  & SSIM$\uparrow$  & 0.815 & 0.758 & 0.851 & 0.839 & 0.821 & 0.877   & 0.845& 0.844  &  \textcolor{red}{0.982}  &  0.979 & \textcolor{blue}{0.981}   \\
  \hline
  \hline
  \multirow{3}{*}{HSTS (Synthetic)}& PSNR$\uparrow$  & 14.48 & 21.57 & 24.48 & 18.64 & 18.92 & 20.55   & 22.94 & 22.06 & \textcolor{blue}{32.75} & 32.72 &  \textcolor{red}{35.48}   \\
  & SSIM$\uparrow$  & 0.761 & 0.873 & 0.915 & 0.817 & 0.741 & 0.897   & 0.874 & 0.847  &  \textcolor{blue}{0.983}  &  0.9781 & \textcolor{red}{0.987}   \\
  \hline
 \end{tabular}
 \label{table1}
\end{table*}

\begin{table}[t]
 \caption{Quantitative (PSNR/SSIM) comparisons with SOTA image dehazing methods on Middlebury~\cite{scharstein2014high} and NH-HAZE~\cite{ancuti2020nh}. The best and second best results are highlighted with \textcolor{red}{red} and \textcolor{blue}{blue} fonts, respectively}
 \centering
 \renewcommand{\arraystretch}{1.4}
 \setlength{\tabcolsep}{0.3mm}
 \begin{tabular}{|c|c|c|c|c|c|c|c|c|}
  \hline
  \multicolumn{2}{|c|}{Method}  & DCP &  AODNet & DCPDN & GFN  & GDN    & DFF   & MSTN (Ours)   \\
  \hline
  \hline
  \multirow{3}{*}{MiddleBury}& PSNR$\uparrow$  & 11.94  & 13.94   & 12.23 & 14.01 & 14.21 & \textcolor{blue}{15.82} &  \textcolor{red}{17.50}   \\
  & SSIM$\uparrow$  & 0.762  & 0.764 & 0.725 & 0.754 & 0.778  &   \textcolor{red}{0.868} & \textcolor{blue}{0.863}   \\
  \hline
  \hline      
  \multirow{3}{*}{NH-HAZE}& PSNR$\uparrow$  & 10.57  & 15.41  & \textcolor{blue}{17.42} & 15.17 & 15.23 & 16.21 &  \textcolor{red}{18.42}   \\
  & SSIM$\uparrow$  & 0.52  & 0.57 & \textcolor{blue}{0.61} & 0.52 & 0.56 & 0.58 & \textcolor{red}{0.63} \\
  \hline
 \end{tabular}
 \label{table-NH}
\end{table}

\section{Experiments}
\subsection{Dataset}
In this paper, we use RESIDE~\cite{li2018benchmarking}, Middlebury~\cite{scharstein2014high}, and NH-HAZE~\cite{ancuti2020nh} to prove the effectiveness of our proposed MSTN on image dehazing task. Moreover, we also adopt the derain dataset (DID-MDN~\cite{zhang2018densityaware}) further verify the effectiveness of the proposed network on other image restoration tasks, thereby verifying the scalability and versatility of MSTN.
\subsubsection{RESIDE}  RESIDE~\cite{li2018benchmarking} is a large-scale image dehazing dataset, which includes synthetic hazy images of indoor and outdoor and real-world hazy images. 
In RESIDE, the atmospheric scattering model is adopted to generate the synthetic hazy images. In this work, we use Indoor Training Set (ITS) and Outdoor Training Set (OTS) as the training dataset and select Synthetic Objective Testing Set (SOTS) and Hybrid Subjective Testing Set (HSTS) as the test dataset, respectively.
Among them, ITS contains 1,399 clear images and 13,990 hazy images with the size of $620 \times 460$. 
Each clear image generates 10 hazy images with $\beta \in[0.6,1.8]$ and $A \in[0.7,1.0]$, and the depth map $d(x)$ comes from the NYU Depth V2~\cite{silberman2012indoor} and Middlebury Stereo datasets~\cite{scharstein2003high}. 
Similar to ITS, OTS also contains a large number of images, but the depth map $d(x)$ of OTS is estimated by using the algorithm developed in~\cite{liu2015learning} and $\beta \in[0.04, 0.2]$, $A \in[0.8, 1.0]$. 
The SOTS contains 500 indoor hazy images and 500 outdoor hazy images, and their generation methods are the same as ITS and OTS, respectively. 
In addition, both synthetic hazy images and real-world hazy images are included in the HSTS.
It is worth noting that the real real-world hazy images in these datasets can be used to verify the dehazing ability of MSTN in real scenes.

\subsubsection{Middlebury Stereo Dataset} Middlebury ~\cite{scharstein2014high} is a high-resolution stereo indoor dataset with subpixel-accurate ground truth. Similar to ITS, the atmospheric scattering model is adopted to generate synthetic hazy images. Considering its high-resolution characteristics, we adopt Middlebury as an assistant testing dataset to demonstrate the robustness of our proposed MSTN. In the experiment, we use the model pre-trained on ITS and directly applied the model to the Middlebury dataset to show the model performance.

\subsubsection{NH-HAZE} NH-HAZE dataset~\cite{ancuti2020nh} was proposed in the NTIRE2020 Image Dehazing Challenge~\cite{ancuti2020ntire}, which is a non-homogeneous realistic dataset that contains 55 outdoor scenes. In NH-HAZE, the haze was be introduced in the scene by using a professional haze generator, which can simulates the real conditions of hazy scenes. Moreover, the hazy and haze-free corresponding scenes contain the same visual content captured under the same illumination parameters. Following the challenge setting, we use images 1 $\sim$ 50 as the training dataset and images 51 $\sim$ 55 as the testing dataset. 

\subsubsection{DID-MDN} DID-MDN~\cite{zhang2018densityaware} is a derain dataset, which includes three different rain-density images, that is light, medium, and heavy rain-density, respectively. In DID-MDN, the training dataset includes 12,000 images and the test dataset includes 12,00 images.

\begin{table}[http]
  \renewcommand{\arraystretch}{1.2}
  \setlength{\tabcolsep}{2mm}
 \caption{Quantitative (SSEQ/BLIINDS-II) comparisons on RESIDE-HSTS~\cite{li2018benchmarking}. The best and second best results are highlighted with \textcolor{red}{red} and \textcolor{blue}{blue} fonts, respectively}
 \centering
 \begin{tabular}{|c|c|c|c|c|}
  \hline
  \multirow{3}{*}{Method}  & \multicolumn{4}{|c|}{HSTS}   \\
  \cline{2-5}
  & \multicolumn{2}{|c|}{Synthetic} & \multicolumn{2}{|c|}{Real} \\
  \cline{2-5} & SSEQ$\downarrow$ & BLIINDS-II$\downarrow$ & SSEQ$\downarrow$ & BLIINDS-II$\downarrow$ \\
  \hline
  \hline
  DCP~\cite{5206515}                      &       86.15 & 90.70   & 68.65 & 69.35 \\
  CAP~\cite{zhu2015fast} &85.32 & 85.75 & 67.67 & 63.55\\
  DehazeNet~\cite{cai2016dehazenet}                &        86.01	&87.15    & 68.34 & 60.35 \\
  MSCNN~\cite{ren2016single} & 85.56 & 88.70 & 68.44 & 60.35 \\
  NLD~\cite{berman2016non} & 86.28& 85.30 & 67.96 & 70.80 \\
  AODNet~\cite{li2017aod}                  &        86.75 & 87.50    & 70.05 & 74.75 \\
  %GFN~\cite{ren2018gated} & 22.94 & 0.874 & - & - \\
  DCPDN~\cite{zhang2018densely}                    &    33.36 & 31.89    & 43.18 & 49.30 \\
  GDN~\cite{liu2019griddehazenet}                      &        \textcolor{blue}{29.59} & \textcolor{blue}{22.89}  & - & - \\
  DFF~\cite{dong2020multi}                      &        31.24 & 25.67    & \textcolor{blue}{37.27} & \textcolor{blue}{34.25} \\ 
  \hline
  MSTN (Ours)              &        \textcolor{red}{28.76} & \textcolor{red}{21.56}    & \textcolor{red}{35.74} & \textcolor{red}{32.55}\\
  \hline
 \end{tabular}
 \label{table-real}
\end{table}
 
\subsection{Implementation Details}
\textbf{Model setting:} In the final version of MSTN, the value of $i$ and $j$ are set as 5, which means that MSTN has 5 branches and the maximum depth of the first branch is 5.
This also means that the model contains 5 branches with different scales.

\textbf{Training setting:}  During training, we use RGB image as
input and augment the image with random rotation($90^{\circ}$, $180^{\circ}$, $270^{\circ}$), horizontal flip, and vertical flip. 
In addition, we randomly crop $240 \times 240$ image patch on the image as the input and set batch-size as 16.
The initial learning rate is $1 \times 10^{-4}$ and cosine annealing strategy~\cite{he2019bag} is applied to adjust the learning rate. 
We implement our model with the PyTorch framework and update it with Adam optimizer ($5\times10^7$ iterations).
All our experiments are performed on GTX TitanXP GPUs.

\begin{figure*}[http]
	\begin{center}
		\includegraphics[width=0.92\linewidth]{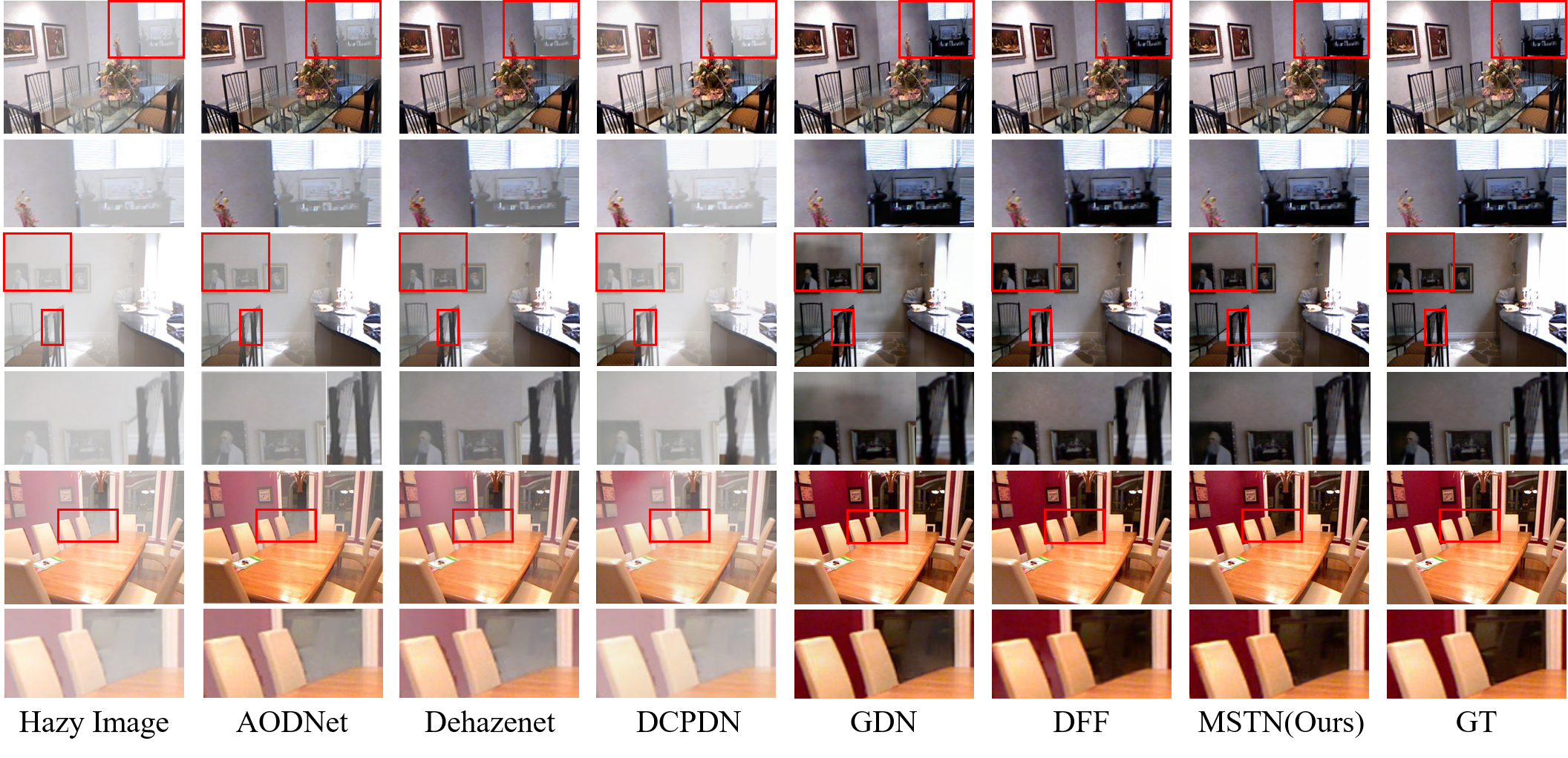}
	\end{center}
	\caption{Visual comparison with SOTA image dehazing methods on the RESIDE-SOTS~\cite{li2018benchmarking} (Indoor) dataset. \textbf{Please zoom in to view details.}}
	\label{indoor}
\end{figure*}

\begin{figure*}[http]
	\begin{center}
		\includegraphics[width=0.92\linewidth]{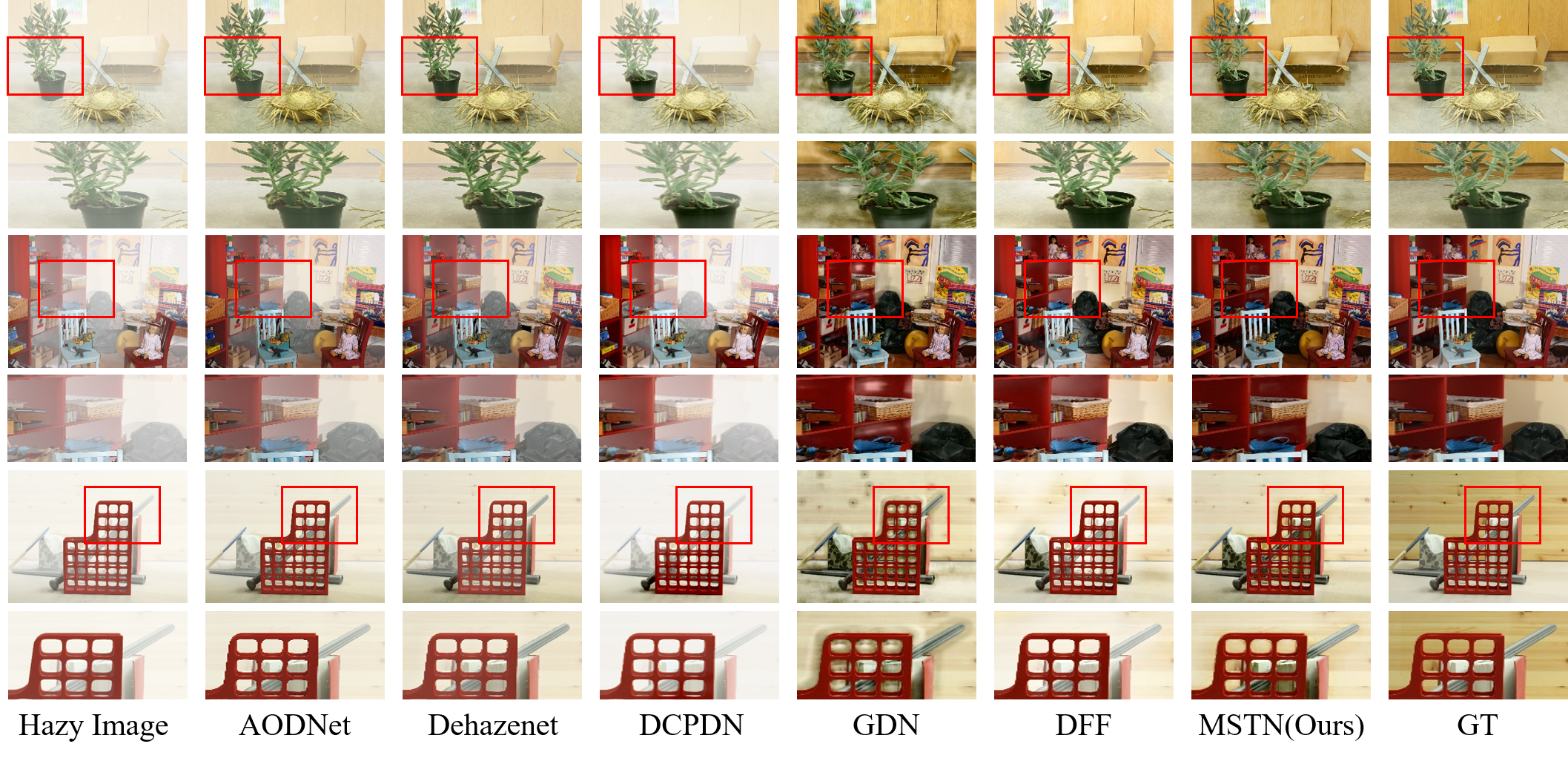}
	\end{center}
	\caption{Visual comparison with the SOTA image dehazing methods on the MiddleBury~\cite{scharstein2014high} dataset. \textbf{Please zoom in to view details.}}
	\label{middle}
\end{figure*}

\begin{figure*}[h]
 \begin{center}
 \includegraphics[width=0.95\linewidth]{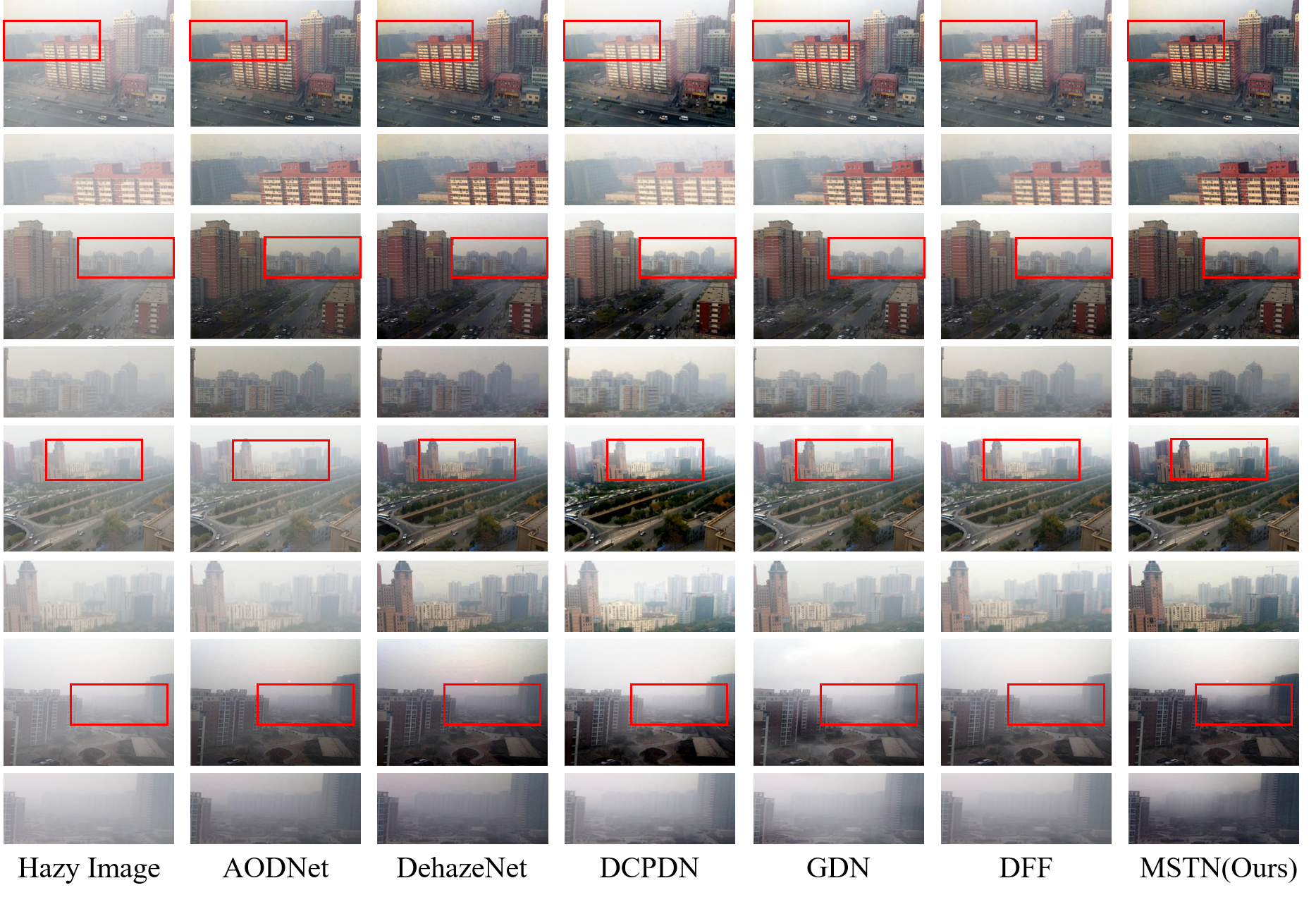}
 \end{center}
 \caption{Visual comparison with SOTA image dehazing methods on the RESIDE-SOTS~\cite{li2018benchmarking} (Outdoor) dataset. \textbf{Please zoom in to view details.}}
 \label{outdoor}
\end{figure*}

\subsection{Comparison with SOTA Image Dehazing Methods}
We compare MSTN with 10 SOTA image dehazing methods, including DCP~\cite{5206515}, CAP~\cite{zhu2015fast}, DehazeNet~\cite{cai2016dehazenet}, MSCNN~\cite{ren2016single}, NLD~\cite{berman2016non}, AODNet~\cite{li2017aod}, DCPDN~\cite{zhang2018densely}, GFN~\cite{ren2018gated}, GDN~\cite{liu2019griddehazenet}, and DFF~\cite{dong2020multi}. 
In addition, we use PSNR, SSIM, SSEQ~\cite{liu2014no}, and BLIINDS-II~\cite{saad2012blind} to evaluate the quality of dehazed images. 
Among them, the larger the PSNR and SSIM value, the better the result.
Contrary, the smaller the SSEQ and BLIINDS-II value, the better the result.
It worth noting that SSEQ and BLIINDS-II are no-reference image quality assessment methods, which can effectively reflect the visual effect of the reconstructed images.

\begin{figure*}[h]
	\begin{center}
		\includegraphics[width=0.92\linewidth]{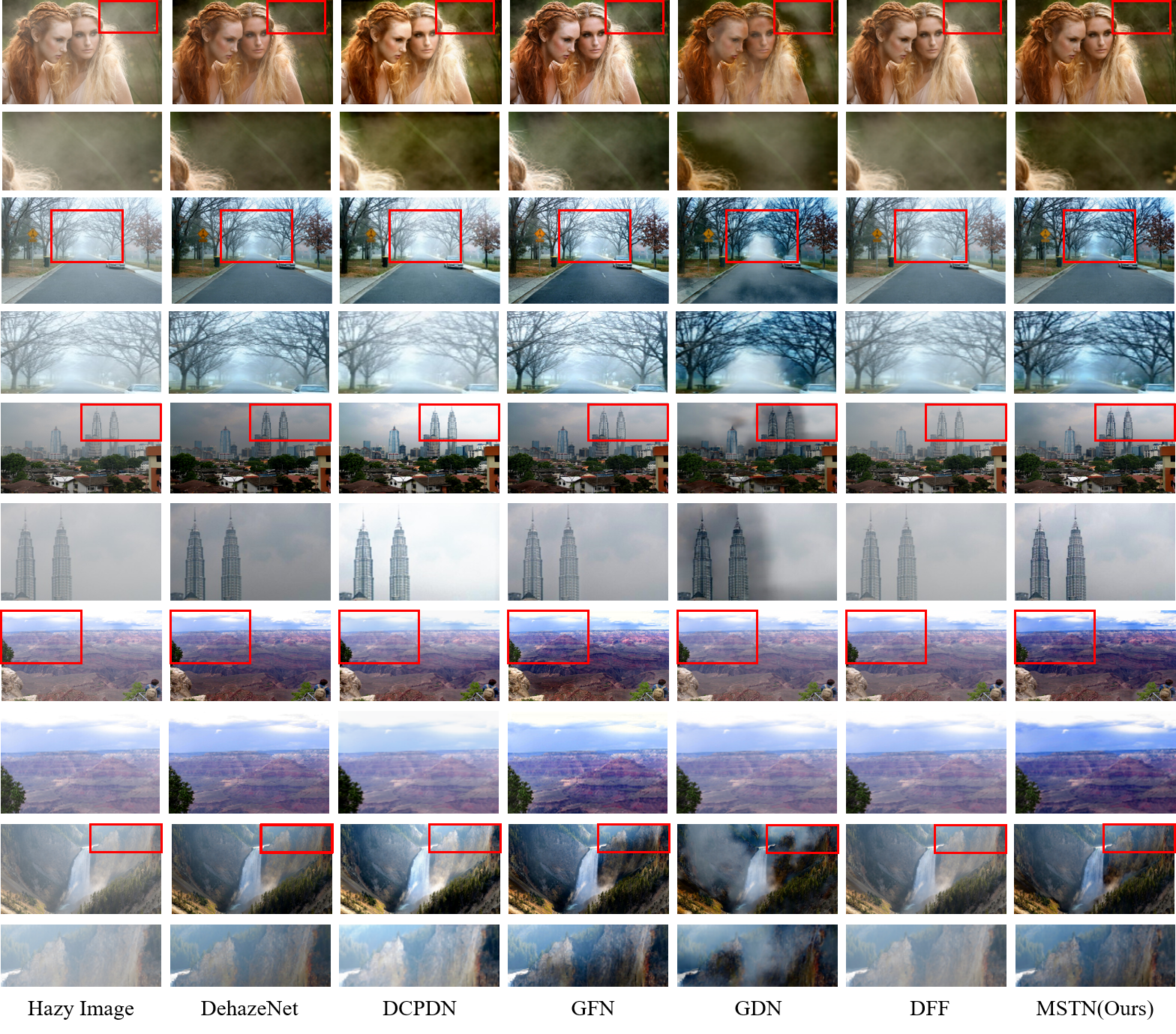}
	\end{center}
	\caption{Visual comparison with SOTA image dehazing methods on real-world hazy images. \textbf{Please zoom in to view details.}}
	\label{Real-world}
\end{figure*}

\subsubsection{\textbf{Quantitative Comparison on Synthesis Images}}
In this work, we trained two different versions of the model for indoor and outdoor scenarios. 
These two models were trained on ITS and OTS datastes, respectively.
In TABLE~\ref{table1}, we provide the PSNR/SSIM comparisons with SOTA image dehaizng methods on SOTS~\cite{li2018benchmarking} (Indoor and Outdoor).
Obviously, our MSTN achieves the best results whether in the indoor or outdoor scenes.
Among these methods, GDN~\cite{liu2019griddehazenet} and DFF~\cite{dong2020multi} are the latest methods and achieved the SOTA results at the time.
Despite this, compared with them, our MSTN still achieved better results with a great advantage.
Specifically, compared to the second-best model, the average PSNR results of MSTN in Indoor and Outdoor scenarios has increased 1.62dB and 0.40dB, respectively.
This is a significant improvement and provide a new SOTA results on the image dehazing task.
This is because the proposed multi-scale topological architecture can extract rich features from the input image, so that the model can reconstruct high-quality haze-free images.
Meanwhile, in order to verify the generalization ability of the model, we directly use the pre-trained MSTN on OTS and ITS to reconstruct haze-free images on HSTS~\cite{li2018benchmarking} (Synthetic) and MiddleBury~\cite{scharstein2014high}, respectively.
According to TABLES~\ref{table1} and~\ref{table-NH}, we can observe that our MSTN still achieves the best results on all of these two datasets.
It is worth mentioning that compared to the second-best model, the results of MSTN on these two datasets are improved by 1.68dB and 2.73dB, respectively.
Moreover, we provide the SSEQ and BLINDS-II comparison of these methods on HSTS~\cite{li2018benchmarking} (Synthetic) in TABLE~\ref{table-real}.
Obviously, our MSTN still achieves the best results.
This further verified the effectiveness and powerful generalization capabilities of MSTN.

\subsubsection{\textbf{Visual Comparison on Synthesis Images}}
In Figs.~\ref{indoor},~\ref{middle}, and~\ref{outdoor}, we show the visual comparison with other image dehazing on SOTS~\cite{li2018benchmarking} (Indoor and Outdoor) and MiddleBury~\cite{scharstein2014high} datasets. 
Among them, the images in the SOTS (Indoor) and MiddleBury datasets contain relatively low haze density while the images in the SOTS (Outdoor) dataset contain high haze concentrations.
According to Figs.~\ref{indoor} and ~\ref{middle}, we can clearly observe that the image reconstructed by AODNet, Dehazenet, and DCPDN still contains a lot of haze.
Compared with these methods, GDN and DFF can reconstruct more clear haze-free images.
However, carefully observing these reconstructed images, we find that these images contain a lot of artifacts and false edges, especially on walls, doors and flat areas.
This greatly limits the promotion and application of these models.
On the contrary, our MSTN can reconstruct high-quality haze-free images without artifacts.
In Fig.~\ref{outdoor}, we show the dehazing effect of the model in the outdoor scenes.
Obviously, outdoor scenes have higher concentrations of haze and the distribution of these haze is uneven.
Therefore, it is more challenging to recover haze-free images from these images.
According to the figure, we can found that all of the compared methods are failed to restoration high-quality images.
Compared with these methods, our MSTN can reconstruct more clear images.
Although the image reconstructed by our MSTN also contains some haze, the performance of MSTN has been greatly improved compared with the previous methods.
This fully demonstrates the effectiveness of MSTN.

\subsubsection{\textbf{Results on Real-World Images}} 
The concentration and distribution of haze in natural scenes are more diverse and complex than the simulated images.
Therefore, the task of real image dehazing is more difficult.
In this part, real hazy image datasets (RESIDE-HSTS~\cite{li2018benchmarking} (Real-world) and NH-HAZE~\cite{ancuti2020nh}) are used to
further assess the practicality of our MSTN.
For HSTS (Real-world), following the setup in RESIDE, we use the model pre-trained on the OTS to reconstruct haze-free images.
For NH-HAZE, we follow the setup of the NTIRE 2020 Challenge on Image Dehazing to train and test our model.
The qualitative results of these two datasets are presented in TABLE~\ref{table-real} (right) and~\ref{table-NH} (bottom), respectively.
According to the results, we can clearly observe that MSTN achieves the best results in all evaluation indicators.
In addition, we provide some dehazing results on real-world images in Fig.~\ref{Real-world}.
According to the figure, we can found that (i). The haze-free images reconstructed by other methods still contains varying degrees of haze; (ii). The haze-free images reconstructed by DCPDN are over-exposed, causing the color of the reconstructed image to deviate; 3) The haze-free images reconstructed by GDN contains a lot of artifacts and the distribution of haze is uneven. 
All of these phenomena expose the flaws of these models.
In contrast, MSTN can reconstruct more clear and realistic haze-free images, which further proves the effectiveness of MSTN in practical applications.

\section{Analysis and Discussion}
\subsection{Study of Model Architecture}
In this article, we propose a Multi-scale Topological Network (MSTN) for image dehazing. In order to study the effectiveness of the proposed architecture,
we provide a series of ablation studies in this section.
It is worth noting that in order to quickly verify the effectiveness of each module, the training settings in this section are as follows: batch size = $8$, patch size = $128 \times 128$, and $1\times10^6$ iterations.

\subsubsection{\textbf{Effectiveness of AFSM}}
\begin{figure}
	\centering
	\includegraphics[scale=0.42]{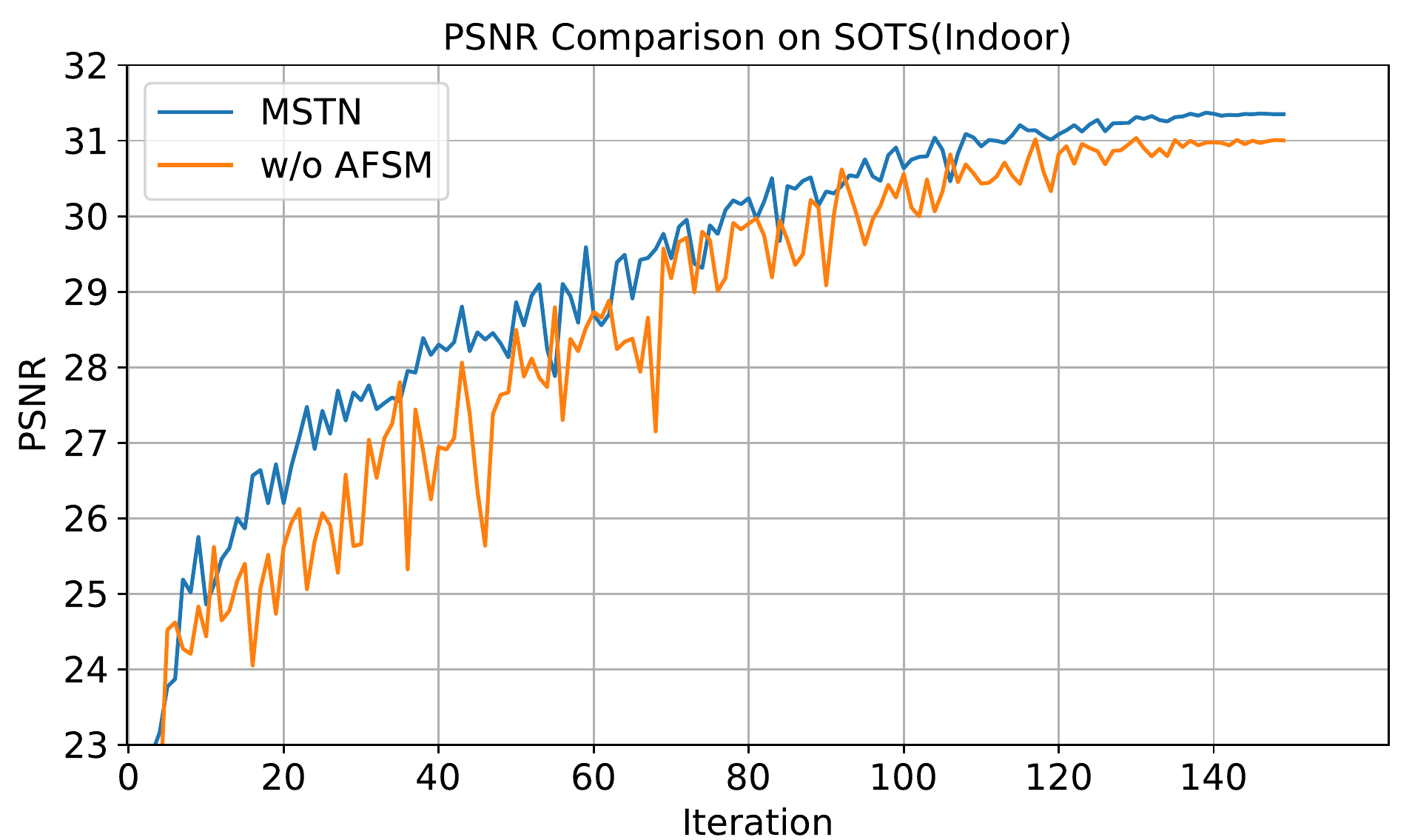}
	\includegraphics[scale=0.42]{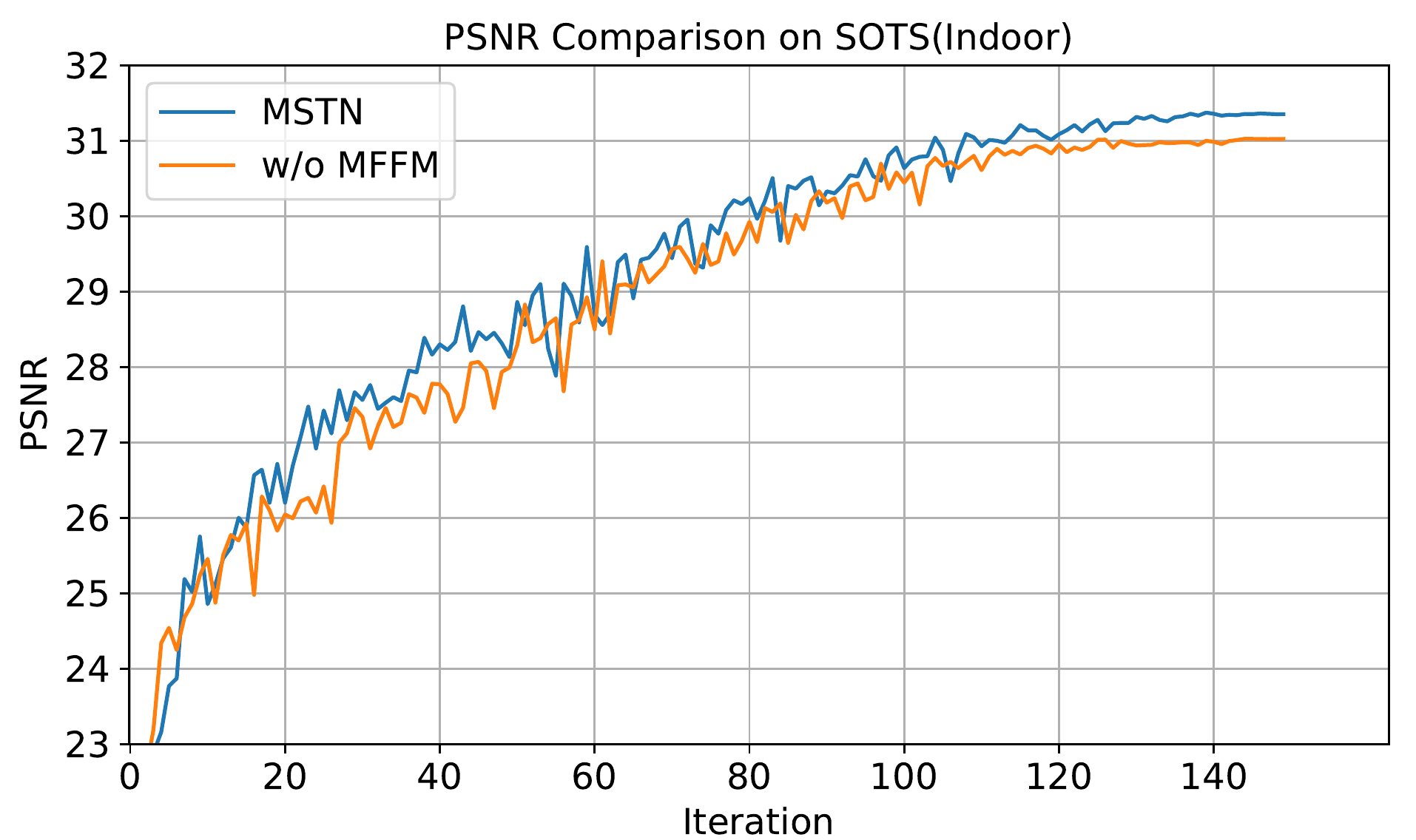}
	\caption{Study on the effectiveness of AFSM and MFFM.}
	\label{mffm}
\end{figure}

AFSM is designed to select and fuse different image features, which also serves as the core component of MFFM.
In order to verify the effectiveness of AFSM, we designed two simplified model, named MSTN (baseline) and MSTN (w/o AFSM).
Among them, MSTN (baseline) was restrained by the new training settings and MSTN (w/o AFSM) has the same architecture with MSTN (baseline) but replaces all AFSMs in the model with the element-wise addition operation.
According to TABLE~\ref{Study-AFSM-MFFM} and Fig.~\ref{mffm}, we can clearly observe that when the element-wise addition operation is used to replace the AFSM, the PSNR result of the model drops by 0.35dB.
Meanwhile, the modified model is unstable during the training, which make the model difficult to converge.

\begin{table}
	\renewcommand{\arraystretch}{1.3}
	 \setlength{\tabcolsep}{1.5mm}
	\caption{Study of AFSM and MFFM on RESIDE-SOTS~\cite{li2018benchmarking} (indoor).}
	\centering
	\begin{tabular}{|c|c|c|c|}
		\hline
		Methods & MSTN (Baseline)  & MSTN (w/o AFSM) & MSTN (w/o MFFM) \\
		\hline
		\hline
		PSNR    & \textbf{31.37}   & 31.02 & 31.03 \\
		SSIM    & \textbf{0.975}   & 0.973 & 0.974 \\
		\hline
	\end{tabular}
	\label{Study-AFSM-MFFM}
\end{table}

\subsubsection{\textbf{Effectiveness of MFFM}}
As we mentioned before, MFFM is the core module of the proposed MSTN, which is designed for multi-scale image feature selection, interaction, and fusion.
According to the cross-scale skip connections, MFFM receives two feature maps with different scales as inputs and output the selected and fused features.
In order to verify the effectiveness of MFFM, we designed a new model, named MSTN (w/o MFFM).
MSTN (w/o MFFM) is a new model that remove all skip connections between different scales and replace all MFFMs in the MSTN (baseline) with residual blocks (RBs).
In TABLE~\ref{Study-AFSM-MFFM} and Fig.~\ref{mffm}, we provide the PSNR results and training curves of MSTN (baseline) and MSTN (w/o MFFM).
According to the results, we can find that when MFFMs are replaced by RBs, the performance of the model drops by 0.34dB.
This greatly illustrates the effectiveness of MFFM.
Meanwhile, this illustrates the importance of multi-scale features and the rationality of MFFM design.

\subsubsection{\textbf{Effectiveness of Multi-scale Architecture}}
\begin{figure}
	\centering
	\includegraphics[scale=0.35]{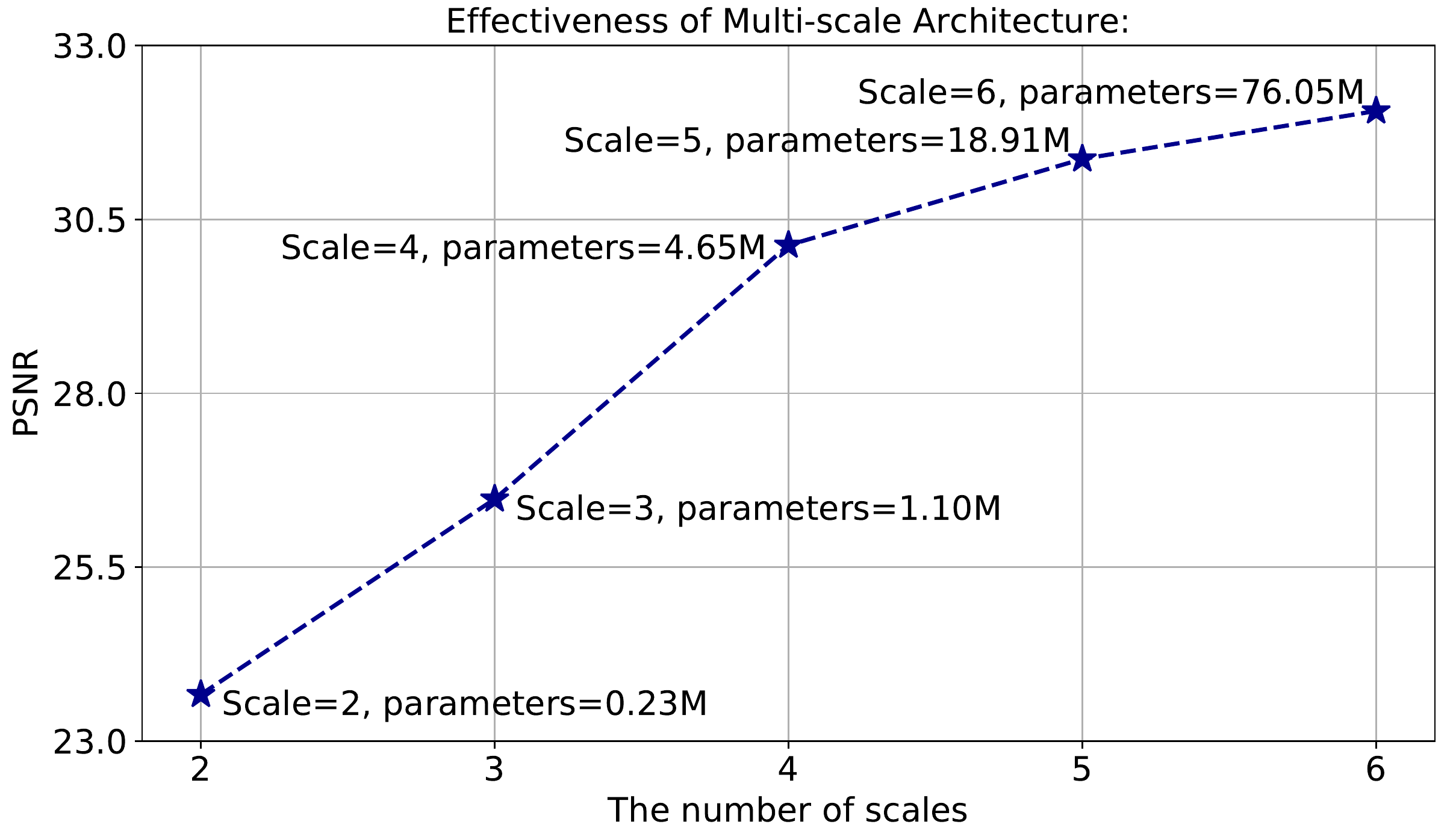}
	\caption{Study the effectiveness of multi-scale architecture on SOTS (indoor).}
	\label{scale-psnr}
\end{figure}

As shown in Fig.~\ref{MSTN}, MSTN adopts the pyramid-like structure to obtained multi-scale image features.
In this paper, the final version of MSTN set $i=5$ and $j=5$.
This means that MSTN can extract image features with 5 different scales.
In order to show the performance of the model under different scales, we designed a new set of models, and set $i=2, j=2$, $i=3, j=3$, $i=4, j=4$, $i=6, j=6$, respectively.
This setting makes these models can extract different number of scales image features.
In Fig.~\ref{scale-psnr}, we show the performance and parameters changes of these models.
Obviously, the PSNR result increases as the scale number increases.
Meanwhile, we can observe that when the number of scales continues to increase (such as $i=6$ and $j=6$), the model performance can be further improved.
This means that the results reported in this paper are not the best results of MSTN.
However, it cannot be ignored that the parameter quantity will increase
as the scale number increases.
Therefore, the number of scale can be selected according to actual demands.
We set $i=5, j=5$ in this paper to achieve a good balance between the model size and performance.

\begin{table}
    \setlength{\tabcolsep}{3.5mm}
	\renewcommand{\arraystretch}{1.3}
	\caption{Study the effectiveness of multi-scale architecture on SOTS (indoor). The best result are highlighted.}
	\centering
	\begin{tabular}{c|c|c|c|c|c}
		\hline
		PSNR & SSIM & Dark gray & Blue & Orange & Gray \\
		\hline
		\hline
		23.05   & 0.881 & $\checkmark$ & & &\\
		27.03   & 0.943 & & $\checkmark$ & &\\
		29.56   & 0.951 & & & $\checkmark$ &  \\
		\textbf{30.45}   & \textbf{0.960} & & & & $\checkmark$ \\
		\hline
	\end{tabular}
	\label{dark}
\end{table}

\begin{table}[b]
    \setlength{\tabcolsep}{4.7mm}
	\renewcommand{\arraystretch}{1.2}
	\caption{Investigations of the model size and execution time.}
	\centering
	\begin{tabular}{ccccc}
		\hline
		\hline
		\textbf{Method}  & \textbf{Param.} & \textbf{Platform} & \textbf{Times (s)}\\
		\hline
		\hline
		DCP     & -      & Matlab       & 1.532 \\
		CAP     & - & matlab &  0.808\\
		DehazeNet & 0.08M & Matlab  & 1.102 \\
		MSCNN     & 0.08& matlab  &  2.48\\
		NLD     & - & matlab  &  9.89\\
		AodNet & 0.02M & Mat-Caffe  & 0.402\\
		GFN & 0.51M & Mat-Caffe  & 1.373\\
		DCPDN &66.89M & Pytorch & 0.248\\
		GDN& 0.96M &Pytorch & 0.0150\\
		DFF & 31.35M &Pytorch & 0.0202\\
		\hline
		Ours & 18.91M &Pytorch & \textbf{0.0139}\\
		\hline
		\hline
	\end{tabular}
	\label{table_time}
\end{table}

As shown in Fig.~\ref{MSTN}, we marked 4 roadmaps (Data Flow) on MSTN with different colors.
This represents four simplified versions of MSTN with different structures.
It is worth noting that, except for the modules marked with "data flow", the modules in these 4 models have been removed.
Meanwhile, these 4 models all have 4 MFFMs.
The only difference between these 4 models is they can extract different types of multi-scale image features.
Specifically, the case 1 model (dark gray data flow) is a flat model, which can only extract image features with one scale.
The case 4 model (gray data flow) is a multi-scale model, which can extract rich multi-scale image features.
According to the TABLE~\ref{dark}, we can clearly observe that when the model can extract more different scales image features, the model can achieve better results.
All these experiments proved the importance of multi-scale image features and the effectiveness of the designed multi-scale architecture.

\subsection{Study of Model Model Size and Execution Time}
\begin{figure}
	\centering
	\includegraphics[scale=0.28]{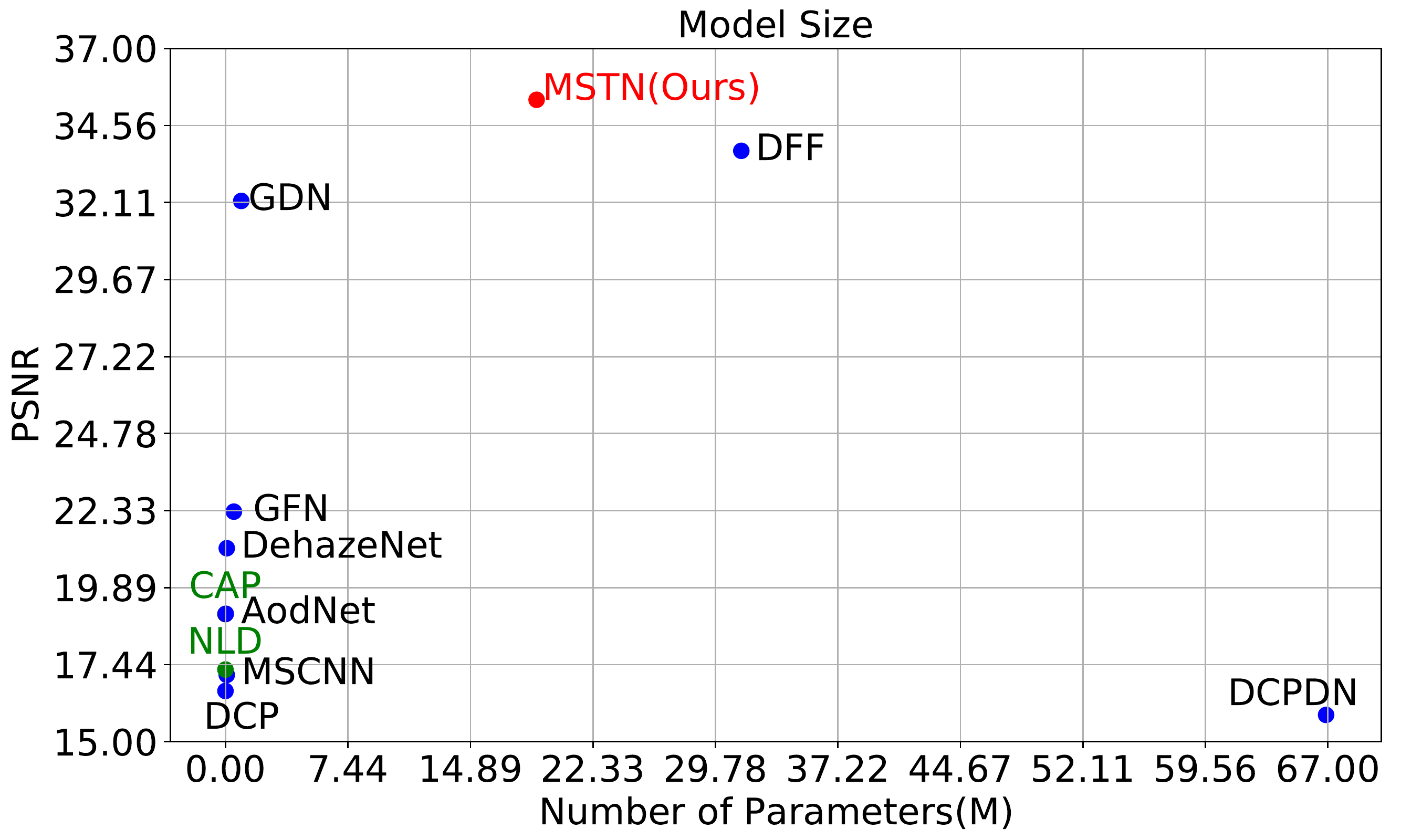}
	\includegraphics[scale=0.28]{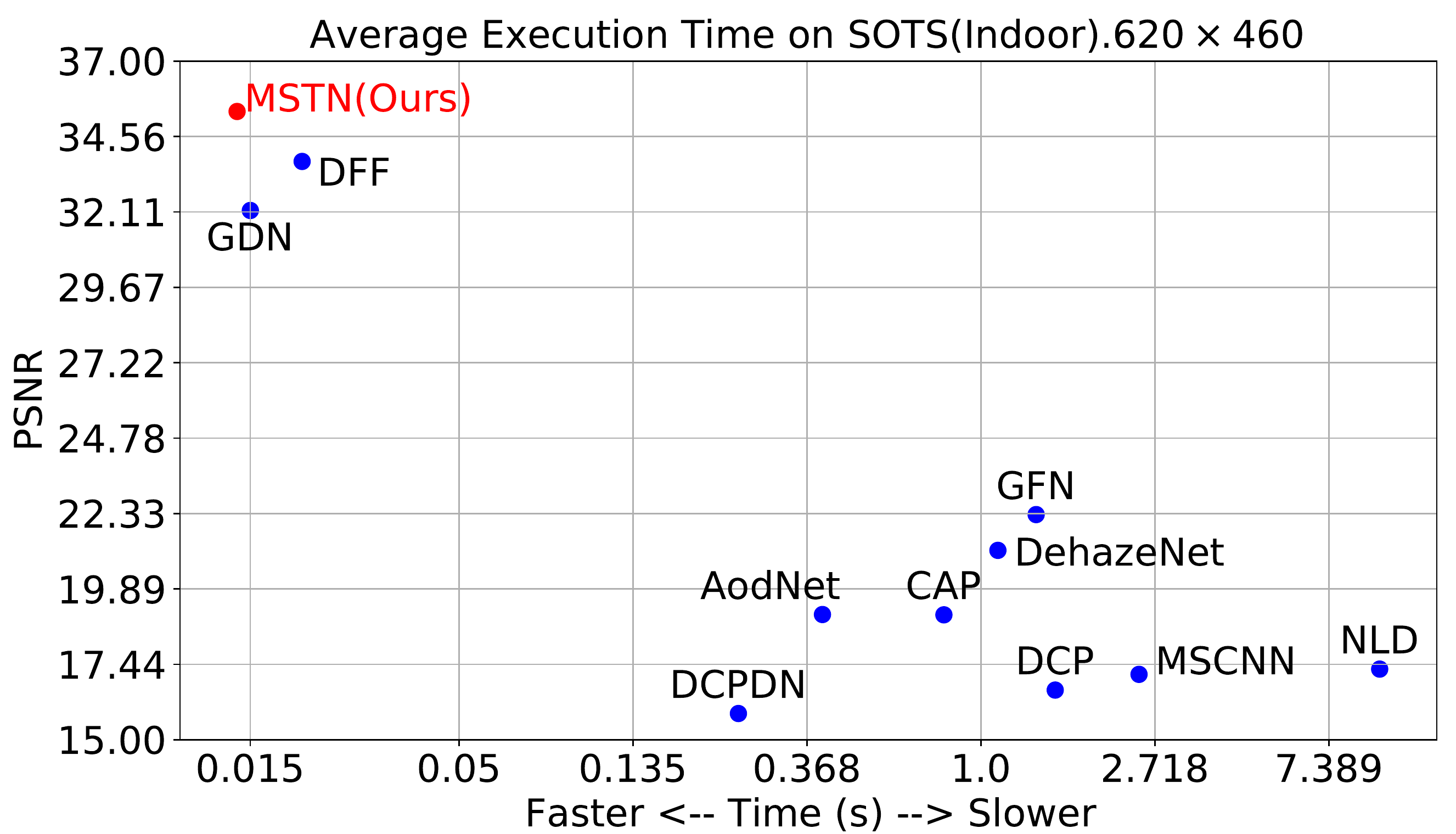}
	\caption{Investigations of the model size and execution time.}
	\label{size}
\end{figure}

Various large size image dehazing models have been proposed in recent years. 
These models always accompanied by numerous parameters, which means that
these models require more storage space, computing resources, and execution time.
In this paper, we aim to explore an efficient and accurate image dehazing model.
Therefore, we need a more efficient network structure, not just increase the model parameters and depth.
In TABLE~\ref{table_time} we show the comparison of model parameters and execution time.
Notice that all repoted models use the released code and test on the same workstation.
The time is the average time required for recovering 500 images of the size of $ 620 \times 460$.
In Fig.~\ref{size}, we intuitively display the comparison of model size, execution time, and performance of each models in the form of dot chart.
According to the figure, we can draw the following conclusions: (1). Compared with lightweight models (e.g., DCP, MSCNN, AODNet, DehazeNet, GFN, GDN), the performance of MSTN is greatly improved; (2). Compared with large models (e.g., DFF and DCPDN), MSTN achieves better results with fewer parameters; (3). Compared with all reported image dehaizng models, MSTN achieves better results with less execution time.
In summary, MSTN achieves a well balance between model performance, size, and
execution time, which provide new solution for real-time image dehazing.

\subsection{Exploring on Other Image Restoration Task}
\begin{figure}
	\centering
	\includegraphics[scale=0.23]{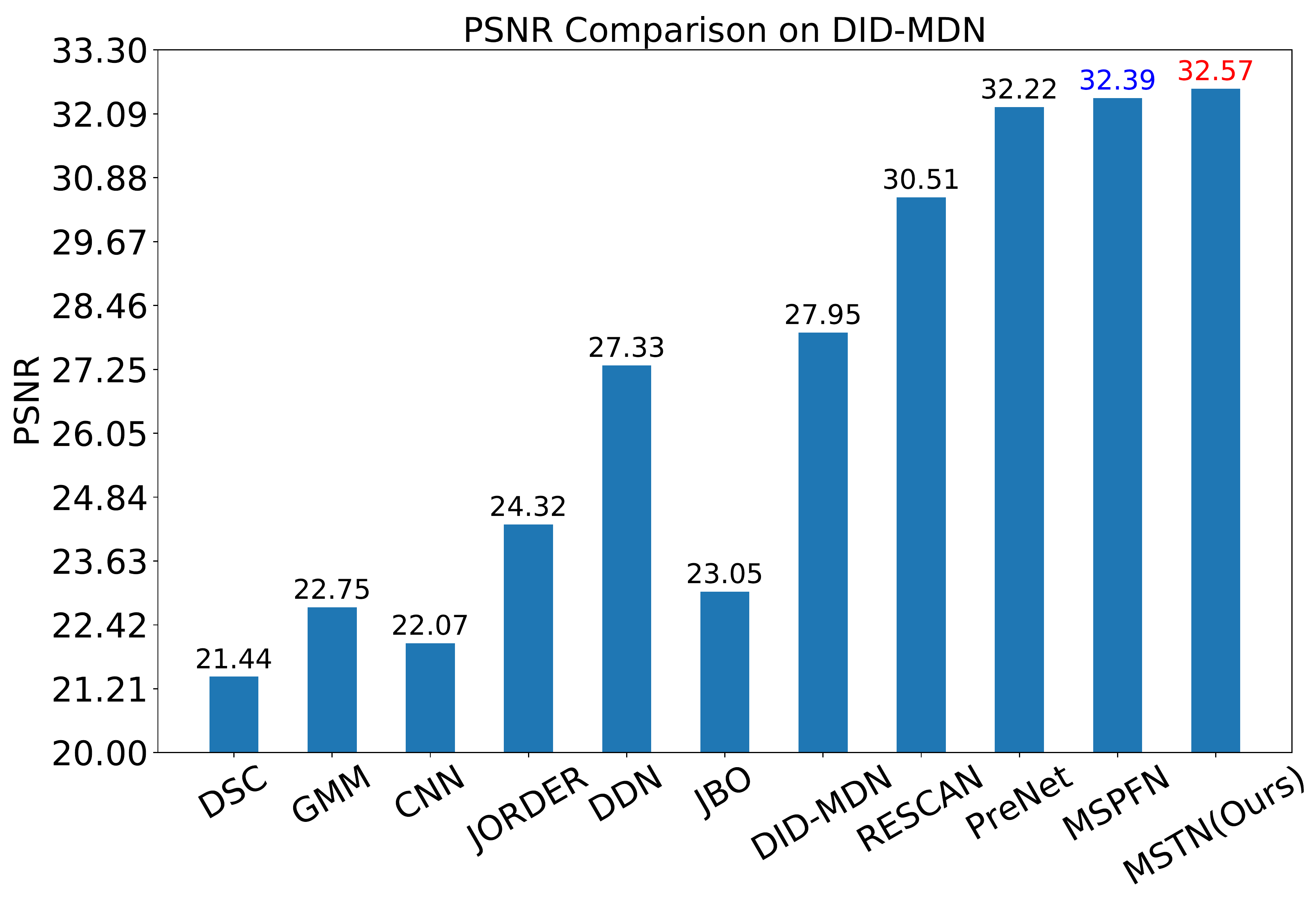}
	\includegraphics[scale=0.23]{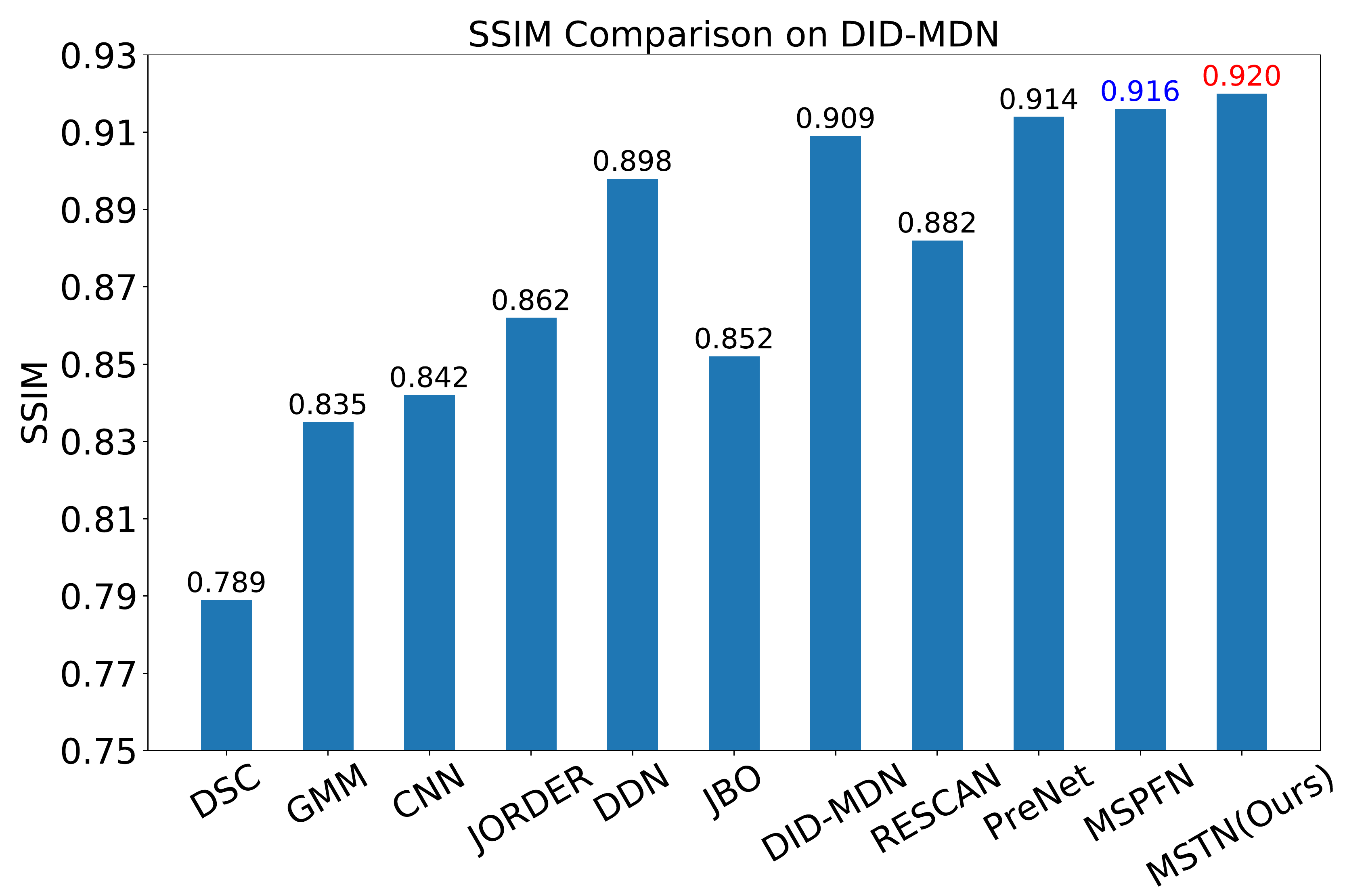}
	\caption{Quantitative comparisons on DID-MDN~\cite{zhang2018densityaware}. The best and second best results are highlighted with \textcolor{red}{red} and \textcolor{blue}{blue} fonts, respectively.}
	\label{did_rain}
\end{figure}

In this paper, MSTN is proposed for the task of single image dehazing.
According to our observation, MSTN is an efficient and accurate multi-scale topological network that can not only suitable for the image dehazing task.
In order to explore the performance of MSTN on other image restoration tasks, we transfer MSTN to the task of single image deraining.
Similar to image dehazing, the task of image deraining aims to reconstruct a clean image from the rain image.
Following previous works, we use DID-MDN~\cite{zhang2018densityaware} to retrain our MSTN and compare it with 10 image deraining models, including DSC~\cite{luo2015removing}, GMM~\cite{li2016rain}, CNN~\cite{fu2017clearing}, JORDER~\cite{yang2017deep}, DDN~\cite{fu2017removing}, JBO~\cite{zhu2017joint}, DID-MDN~\cite{zhang2018densityaware}, RESCAN~\cite{li2018recurrent}, PreNet~\cite{ren2019progressive}, and MSPFN~\cite{jiang2020multi}. 
PSNR and SSIM results are provide in Fig.~\ref{did_rain}.
According to the figure, we can clearly observe that MSTN achieves the best results in both PSNR and SSIM.
This further proves the effectiveness of MSTN.
This also means that MSTN is a highly scalable model that can be applied to other image restoration tasks.
In future works, we will further verify the versatility and robustness of MSTN on other image restoration tasks like image desnowing and image denoising.

\begin{figure*}
	\centering
	\includegraphics[scale=0.56]{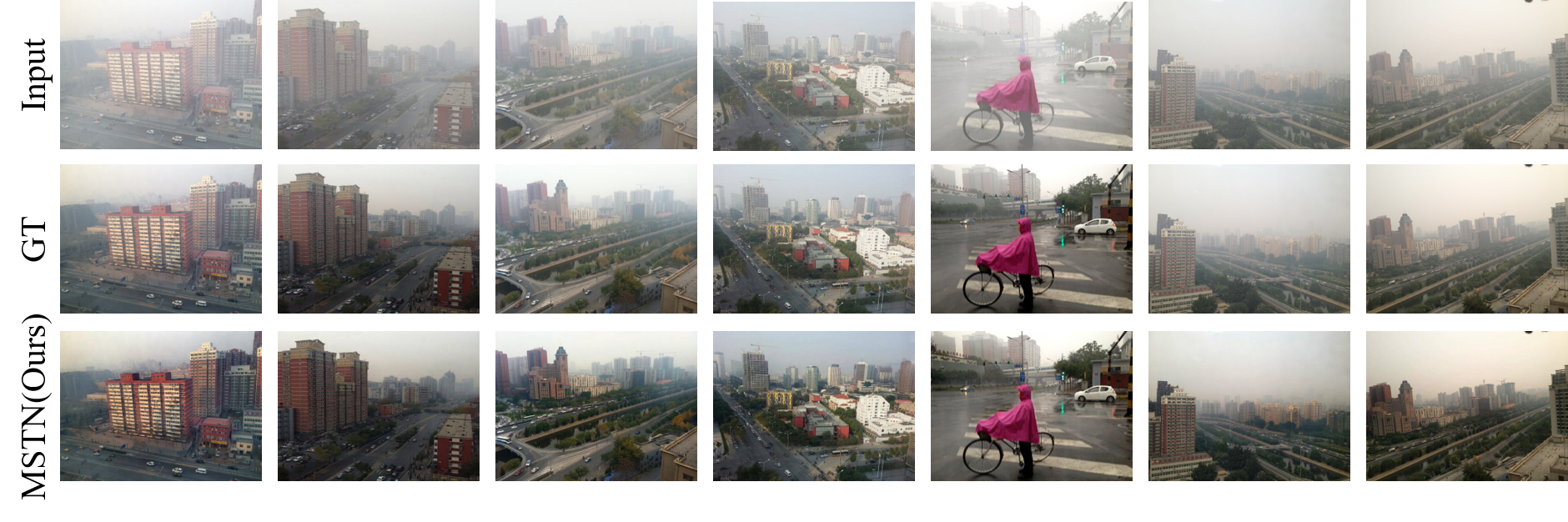}
	\caption{Study on hazy images (RESIDE-SOTS~\cite{li2018benchmarking} (Outdoor)). Obviously, MSTN can reconstruct clear and high-quality haze-free images.}
	\label{real}
\end{figure*}

\subsection{Study on Hazy Images}
In Fig.~\ref{outdoor}, we provide the image reconstructed by MSTN on the RESIDE-SOTS~\cite{li2018benchmarking} (Outdoor) dataset.
According to the figure, we can clearly observe that MSTN can reconstruct more clear images compared to other models.
Moreover, we found that the image reconstructed by our MSTN is even clearer than the GT (Ground-Truth) image (Fig.~\ref{real}).
This phenomenon attracted our attention.
Therefore, we re-investigated the RESIDE-SOTS~\cite{li2018benchmarking} (Outdoor) dataset.
This dataset contains 500 indoor hazy iamges and all these images are synthetic image.
We investigated these images and found that these images contain different concentrations of haze itself. 
Therefore, part of the GT images in this dataset are hazy.
However, there are plenty of clear GT images in the training dataset, thus the powerful learning ability of MSTN can learn how to reconstruct haze-free images from the hazy image.
Therefore, our MSTN can reconstruct clearer images than GT images.
This is because the dehazing ability learned by MSTN can remove the haze from the GT image itself.
In Fig.~\ref{Real-world}, we provide the reconstruction results of MSTN on real hazy images.
Obviously, MSTN can reconstruct high-quality haze-free images on real hazy images.
This further proves the effectiveness and practicality of MSTN.

\section{Conclusions}
In this paper, we proposed an efficient and accurate Multi-scale Topological Network (MSTN) for single image dehaizng, which achieved competitive results on multiple datasets.
MSTN adopts a new type of multi-scale topological architecture, which provides a large number of search paths and topological sub-networks that can fully extract image features from the input hazy image and improve the model stability and robustness.
Meanwhile, we proposed a Multi-scale Feature Fusion Module (MFFM) and an Adaptive Feature Selection Module (AFSM) to realize the automatic transmission, selection, and fusion of multi-scale image features.
Extensive experiments show that this special structure makes our model can extract rich image features to reconstruct high-quality haze-free images with texture details. 
Additionally, we achieved promising results by applying the model to other image restoration tasks such as image deraining.
This further proves the effectiveness and versatility of the model.
In future works, we will further verify the performance of the proposed model in more image restoration tasks.
% use section* for acknowledgment

% Can use something like this to put references on a page
% by themselves when using endfloat and the captionsoff option.
\ifCLASSOPTIONcaptionsoff
  \newpage
\fi

% trigger a \newpage just before the given reference
% number - used to balance the columns on the last page
% adjust value as needed - may need to be readjusted if
% the document is modified later
%\IEEEtriggeratref{8}
% The "triggered" command can be changed if desired:
%\IEEEtriggercmd{\enlargethispage{-5in}}

% references section

% can use a bibliography generated by BibTeX as a .bbl file
% BibTeX documentation can be easily obtained at:
% http://mirror.ctan.org/biblio/bibtex/contrib/doc/
% The IEEEtran BibTeX style support page is at:
% http://www.michaelshell.org/tex/ieeetran/bibtex/
%\bibliographystyle{IEEEtran}
% argument is your BibTeX string definitions and bibliography database(s)
%\bibliography{IEEEabrv,../bib/paper}
%
% <OR> manually copy in the resultant .bbl file
% set second argument of \begin to the number of references
% (used to reserve space for the reference number labels box)
\bibliographystyle{IEEEtran}
\bibliography{dehaze}

% biography section
% 
% If you have an EPS/PDF photo (graphicx package needed) extra braces are
% needed around the contents of the optional argument to biography to prevent
% the LaTeX parser from getting confused when it sees the complicated
% \includegraphics command within an optional argument. (You could create
% your own custom macro containing the \includegraphics command to make things
% simpler here.)
%\begin{IEEEbiography}[{\includegraphics[width=1in,height=1.25in,clip,keepaspectratio]{mshell}}]{Michael Shell}
% or if you just want to reserve a space for a photo:

% You can push biographies down or up by placing
% a \vfill before or after them. The appropriate
% use of \vfill depends on what kind of text is
% on the last page and whether or not the columns
% are being equalized.

%\vfill

% Can be used to pull up biographies so that the bottom of the last one
% is flush with the other column.
%\enlargethispage{-5in}

% that's all folks
\end{document}